%% file: main.tex
\documentclass{article} %
\usepackage{colm2024_conference}

\usepackage{booktabs}
\usepackage{graphicx}
\usepackage{enumitem}
\usepackage{wrapfig}
\usepackage{algorithm}
\usepackage{algpseudocode}
\usepackage{wrapfig}
\usepackage{float}
\usepackage{microtype}
\usepackage{amsmath}
\usepackage{amssymb}
\usepackage{colortbl}
\usepackage[utf8]{inputenc}
\usepackage{caption}
\usepackage{subcaption}
\usepackage{xcolor}
\usepackage{setspace}
\usepackage{url}
\usepackage{multirow}
\usepackage{colortbl}
\usepackage{tabularx}
\usepackage{blindtext}
\usepackage{pgfplots}
\pgfplotsset{compat=1.18}
\usepackage{tikz}
\usetikzlibrary{er, positioning, bayesnet}
\usepackage{makecell}
\usepackage{tipa}
\usepackage{siunitx}
\usepackage{nicefrac}
\usepackage{tocloft}
\usepackage{listings}
\usepackage[raster, skins]{tcolorbox} %
\usepackage{xltabular}
\usepackage{adjustbox}
\usepackage{xurl}
\usepackage{longtable}  
\input{math_commands.tex}

\definecolor{lightgray}{rgb}{0.9,0.9,0.9}

\renewcommand{\arraystretch}{1.25}

\makeatletter
\DeclareRobustCommand\onedot{\futurelet\@let@token\@onedot}
\def\@onedot{\ifx\@let@token.\else.\null\fi\xspace}

\def\eg{\emph{e.g}\onedot}

\makeatother

\title{
PromptEnhancer:\\ A Simple Approach to Enhance Text-to-Image Models \\via Chain-of-Thought Prompt Rewriting
}

\author{ \bf Tencent Hunyuan }

\newcommand{\basemodel}{HunyuanImage 2.1\xspace}
\newcommand{\methodname}{PromptEnhancer\xspace}
\newcommand{\methodnamebf}{\textbf{PromptEnhancer}\xspace}

\begin{document}

\maketitle

\input{sec/0_abstract}
\input{sec/1_introduction}
\input{sec/2_methodology}
\input{sec/3_0_data_pipe}
\input{sec/3_experiment}
\input{sec/4_relatedwork}
\input{sec/5_conclusion}

\clearpage
\input{sec/x_contributor}
\clearpage

\bibliography{colm2024_conference}
\bibliographystyle{colm2024_conference}
\end{document}

%% file: math_commands.tex
\usepackage{amsmath,amsfonts,bm}

\def\eqref#1{equation~\ref{#1}}

\def\1{\bm{1}}

\DeclareMathAlphabet{\mathsfit}{\encodingdefault}{\sfdefault}{m}{sl}
\SetMathAlphabet{\mathsfit}{bold}{\encodingdefault}{\sfdefault}{bx}{n}



%% file: sec/0_abstract.tex
\begin{abstract}
Recent advancements in text-to-image (T2I) diffusion models have demonstrated remarkable capabilities in generating high-fidelity images. However, these models often struggle to faithfully render complex user prompts, particularly in aspects like attribute binding, negation, and compositional relationships. 
This leads to a significant mismatch between user intent and the generated output. To address this challenge, we introduce \methodnamebf, a novel and universal prompt rewriting framework that enhances any pretrained T2I model without requiring modifications to its weights. 
Unlike prior methods that rely on model-specific fine-tuning or implicit reward signals like image-reward scores, our framework decouples the rewriter from the generator. 
We achieve this by training a Chain-of-Thought (CoT) rewriter through reinforcement learning, guided by a dedicated reward model we term the \textbf{AlignEvaluator}.
The AlignEvaluator is trained to provide explicit and fine-grained feedback based on a systematic taxonomy of 24 key points, which are derived from a comprehensive analysis of common T2I failure modes. By optimizing the CoT rewriter to maximize the reward from our AlignEvaluator, our framework learns to generate prompts that are more precisely interpreted by T2I models. Extensive experiments on the \basemodel model demonstrate that \methodname significantly improves image-text alignment across a wide range of semantic and compositional challenges. 
Furthermore, we introduce a new, high-quality human preference benchmark to facilitate future research in this direction.
\end{abstract}

%% file: sec/1_introduction.tex
\section{Introduction}
The proliferation of large-scale text-to-image (T2I) diffusion models~\citep{ddpm_ho_2020,DALLE_ramesh_2021,imagen_saharia_2022,LDM_rombach_2022,dit_peebles_2023,sd35_esser_2024,gpt4o_hurst_2024}, such as Imagen~\citep{imagen_saharia_2022}, Stable Diffusion~\citep{LDM_rombach_2022,sdxl_podell_2024,sd35_esser_2024}, HunyuanDiT~\citep{hunyuandit_li_2024}, Flux~\citep{Flux_blacklabs_2025}, and Qwen-Image~\citep{QwenImage_wu_2025}, has revolutionized content creation by enabling the synthesis of diverse and photorealistic images from natural language. 
Despite their impressive capabilities, the quality and accuracy of the generated output are heavily contingent on the user's ability to craft precise and detailed prompts. A persistent challenge lies in the inherent ambiguity and conciseness of typical user inputs, which often causes T2I models to misinterpret crucial details, such as failing to bind attributes to the correct objects, ignoring negations, or struggling with complex spatial relationships. This discrepancy highlights a fundamental gap between human intention and model interpretation.

To bridge this gap, prior research has explored various prompt rewriting techniques~\citep{reprompt_wu_2025, hao2023optimizing}. These methods aim to automatically enrich a user's initial prompt to provide more explicit guidance to the T2I model. However, a major limitation of existing approaches is their lack of universality. Many rewriters require model-specific fine-tuning or are co-trained with a particular generator. 
Others rely on optimizing for generic, implicit reward signals such as CLIP scores~\citep{CLIP_radford_2021} or general human preference scores~\citep{imagereward_xu_2023,HPS_Wu_2023}. 
These rewards offer only a coarse measure of overall image-text similarity and lack the specificity to diagnose and correct fine-grained errors. This dependency on specific T2I architectures and ambiguous reward functions renders them difficult to adapt to new models and computationally expensive, restricting their practical applicability in the rapidly evolving landscape of generative models.
\begin{figure}[th]
\centering
\includegraphics[width=1.0\textwidth]{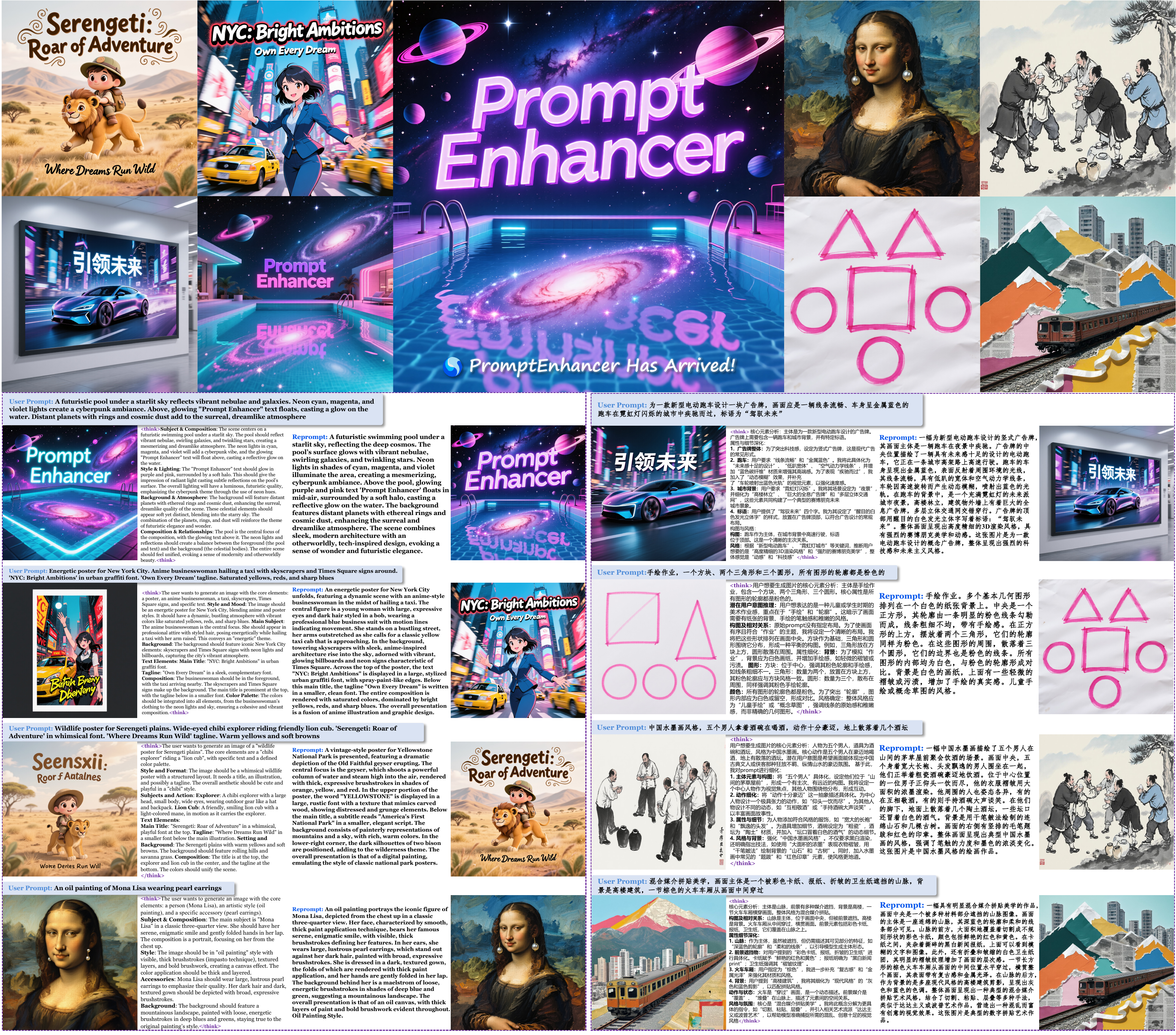}
\caption{
\textbf{\methodname enables high-fidelity and stylistically diverse image generation from user prompts.}
Using \basemodel as the base T2I model, our method demonstrates its versatility across various domains, including photorealism, digital art, abstract geometry, and multilingual text-in-image generation. 
The examples showcase how minimal user inputs are transformed into rich, detailed prompts that yield high-quality visual outputs, bridging the gap between user intent and model execution.
}
\end{figure}

In this paper, we propose \methodnamebf, a universal and model-agnostic prompt rewriting framework designed to enhance any pretrained T2I model without the need for retraining or architectural changes. 
The core idea is to completely decouple the task of prompt refinement from the image generation process.
Our framework introduces a sophisticated rewriter that employs a Chain-of-Thought (CoT) method~\citep{CoT_wei_2022}. 
This approach emulates a human-like reasoning process, allowing the rewriter to systematically deconstruct the initial prompt into its core semantic components, identify potential ambiguities, and then enrich it with explicit details concerning object attributes, spatial arrangements, and complex interactions. 
The resulting detailed prompt can then be used by a T2I model to generate a more faithful image.

The key to our framework's effectiveness lies in how the CoT rewriter is trained. We introduce a novel reward model, the \textbf{AlignEvaluator}, which is specifically designed for fine-grained, multi-faceted evaluation. Instead of relying on a single, holistic score, the AlignEvaluator assesses a generated image against a comprehensive set of 24 distinct key points. These points are meticulously organized into six distinct categories, covering a wide spectrum of challenges including linguistic understanding (\eg, negation, pronoun reference), visual attribute accuracy, compositional relationships, and world knowledge. By optimizing the CoT rewriter to generate prompts that maximize the explicit, structured reward from the AlignEvaluator, we effectively align the prompt generation process with a nuanced understanding of human preferences. 
This ensures that the rewritten prompts are not merely longer, but are structured in a way that elicits higher-quality and more accurate results from the downstream T2I model.

Our main contributions are as follows:

\begin{itemize}[leftmargin=2em,itemsep=1pt,topsep=0pt,partopsep=0pt,parsep=2pt]
\item We propose \methodnamebf, a novel and universal prompt rewriting framework. It introduces a Chain-of-Thought (CoT) mechanism to systematically deconstruct and enrich user prompts, significantly improving the performance of existing T2I models without requiring any modification to their weights.
\item We develop a multi-faceted reward model, the AlignEvaluator, trained on 24 distinct key points organized into 6 categories, enabling fine-grained evaluation and effective alignment of the rewriter.
\item We provide extensive empirical evidence of our method's effectiveness on a state-of-the-art T2I model and contribute a new, high-quality benchmark based on human evaluation to the community.
\end{itemize}

%% file: sec/2_methodology.tex
\section{Methodology}
\label{sec:method}
\begin{figure}[h]
\centering
\includegraphics[width=1.0\linewidth]{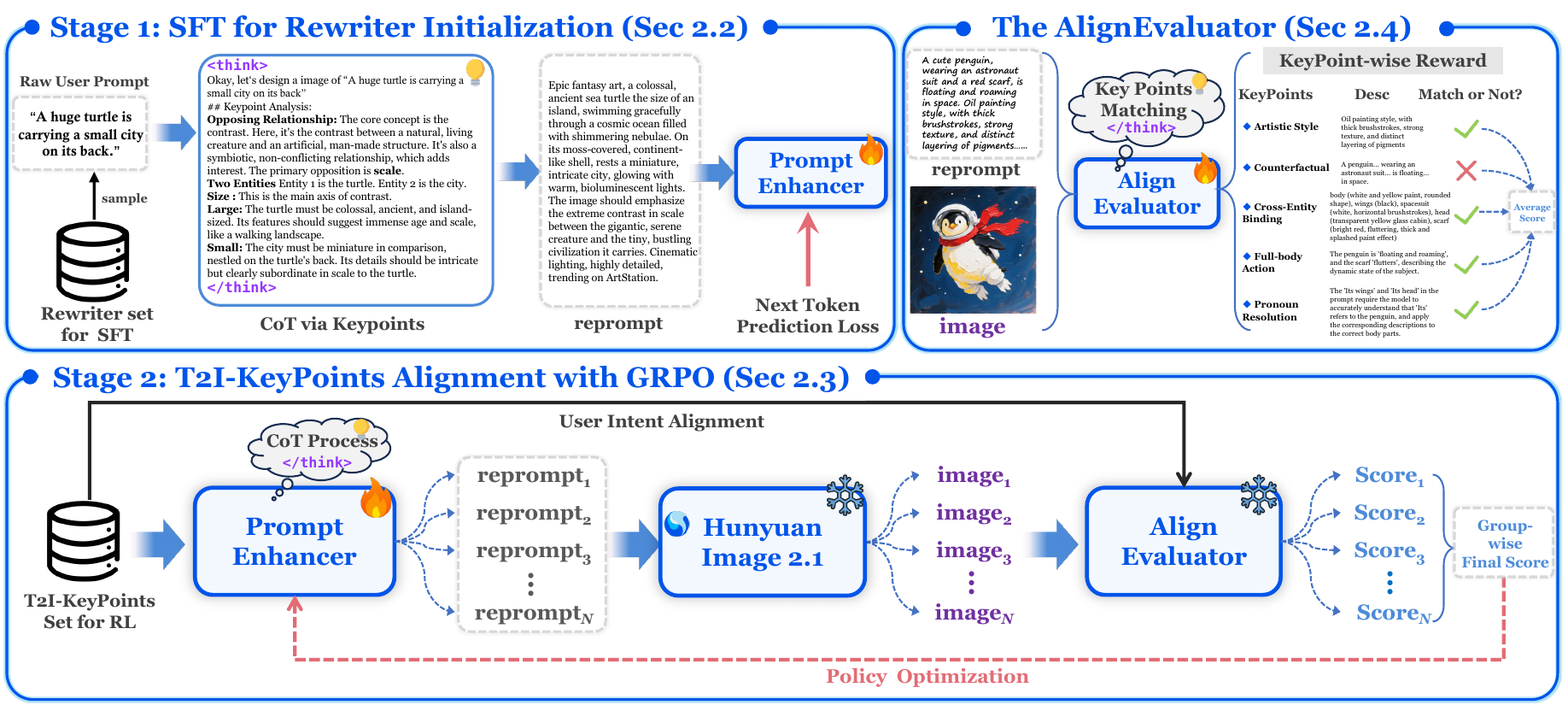}
\vspace{-1.5em}
\caption{
\textbf{
An overview of the training framework for \methodname.
}
Our framework trains a universal Rewriter to enhance pretrained Text-to-Image (T2I) model without altering its weights. 
This is achieved through a two-stage process guided by a specialized reward model.
\textbf{Stage 1: SFT for Rewriter Initialization} (Sec~\ref{sec:stage1_sft4cot}). 
The CoT Rewriter is first initialized via SFT on (user prompt, reprompt) pairs. This stage teaches the model to generate structured, chain-of-thought style responses using a standard next-token prediction loss, establishing a strong foundation for refinement.
\textbf{Stage 2: Policy Alignment with GRPO} (Sec~\ref{sec:stage2_grpo}). The initialized rewriter is further refined using GRPO. 
The rewriter generates multiple reprompt candidates, which are used by a frozen T2I model to create images. The pre-trained AlignEvaluator then assesses each (image, user prompt) pair and provides a scalar reward. 
This reward signal optimizes the rewriter's policy, steering it toward generating prompts that maximize the alignment between the image and the user's intent.
\textbf{The AlignEvaluator} (Sec~\ref{sec:alignevaluator}). Central to our framework is the AlignEvaluator, a pre-trained reward model. It is trained on a large-scale dataset annotated against a taxonomy of 24 fine-grained key points (T2I-KeyPoints, Tab.~\ref{tab:keypoints}). This enables it to provide a robust and nuanced reward signal, which is crucial for the policy alignment stage.
} \label{fig:pipeline}
\end{figure}

In this section, we delineate the methodology of \methodname. 
Our framework is centered around two core components: a \textbf{CoT Rewriter}, which enriches user prompts into detailed instructions, and an \textbf{AlignEvaluator}, a reward model that provides fine-grained feedback on image-text alignment. 
The primary objective is to train the CoT Rewriter through a two-stage pipeline. 
\textit{First}, we initialize the rewriter via Supervised Fine-Tuning (SFT, Sec.~\ref{sec:stage1_sft4cot}) to master the Chain-of-Thought format. 
\textit{Second}, we refine it using Group Relative Policy Optimization (GRPO, Sec.~\ref{sec:stage2_grpo})~\citep{deepseekmath_shao_2024}, where the AlignEvaluator (Sec.~\ref{sec:alignevaluator}) guides the rewriter to generate prompts that are optimally understood by downstream Text-to-Image (T2I) models.

\subsection{Overall Framework}
\label{sec:overall_framework}

As illustrated in Figure~\ref{fig:pipeline}, the \methodnamebf framework consists of three main components:
\begin{itemize}[leftmargin=2em,itemsep=1pt,topsep=0pt,partopsep=0pt,parsep=2pt]
\item \textbf{The CoT Rewriter a.k.a \methodname}: A policy model based on a large vision-language model, which is responsible for reformulating an initial user prompt into an enriched one.
\item \textbf{The AlignEvaluator}: A reward model that assesses a generated (image, prompt) pair based on 24 fine-grained key points and outputs a scalar reward signal.
\end{itemize}

\textbf{Training Objective.} The entire training process aims to optimize the CoT-based Rewriter. 
It first undergoes SFT for basic instruction following and then is refined using feedback from the AlignEvaluator in a reinforcement learning loop.

\subsection{Stage 1: Supervised Fine-Tuning for Rewriter Initialization}
\label{sec:stage1_sft4cot}
\textbf{Objective of SFT.} The first stage of our pipeline is Supervised Fine-Tuning, which equips the CoT-based Rewriter with the fundamental ability to generate structured, chain-of-thought style responses. 
The objective is not to perfect the rewriter, but to provide it with a strong initialization point for the subsequent alignment stage. 

\textbf{Instruction Data Distillation.} To achieve this, we construct a high-quality dataset of (user prompt, reprompt) pairs by leveraging a powerful proprietary large model (\eg, Gemini-2.5-Pro~\citep{gemini2.5_comanici_2025} for English and DeepSeekV3~\citep{deepseekv3_liu_2024} for Chinese) as a ``teacher'' for distillation. For a given short user prompt, the teacher model produces a detailed breakdown of the prompt's elements, analyzes potential ambiguities, and finally synthesizes a comprehensive, enriched prompt. 
The CoT Rewriter is then fine-tuned on this dataset using a standard language modeling objective.

\subsection{Stage 2: Policy Alignment with GRPO}
\label{sec:stage2_grpo}
\textbf{Policy Alignment via GRPO.} After SFT, the rewriter can generate plausible long prompts, but they are not explicitly optimized for image-text alignment. 
The second stage therefore employs Group Relative Policy Optimization (GRPO~\citep{deepseekmath_shao_2024}) to align the user prompt with the fine-grained preferences captured by our AlignEvaluator. 

\textbf{The GRPO Training Loop.} This training process is iterative. For a given user prompts $\{p_1, p_2, ..., p_N\}$, the CoT Rewriter first generates $N$ candidate rewritten prompts $\{p'_1, p'_2, ..., p'_N\}$. 
Each of these prompts is then fed into the frozen T2I model to generate a corresponding image $I_i$. Subsequently, our AlignEvaluator calculates a scalar reward $r_i$ for each pair $(p_i, I_i)$.
Finally, these rewards $\{r_1, r_2, ..., r_N\}$ are used to update the rewriter's policy, creating a preference ranking that encourages the generation of prompts leading to higher rewards.

\subsection{The AlignEvaluator: A Multi-faceted Reward Model}
\label{sec:alignevaluator}
\textbf{Motivation for Fine-grained Reward.} A generic reward signal like a CLIP score~\citep{CLIP_radford_2021} is insufficient for our goal, as it fails to capture nuanced aspects of prompt following. 
We therefore developed the AlignEvaluator, a reward model trained specifically to assess image-text alignment across a wide array of challenges. 

\textbf{Taxonomy of T2I-KeyPoints.} The evaluator is built upon a taxonomy of 24 fine-grained key points, organized into six major categories. These categories cover a comprehensive spectrum of abilities, including: Linguistic and Grammatical Understanding, which assesses the interpretation of core language structures like \textit{Negation}; Visual Attributes, which focuses on rendering properties like \textit{Object Count}; Actions and Interactions, which measures the depiction of dynamic states; Relational and Compositional Structure, which evaluates the handling of complex scenes; and higher-level abilities covered by Knowledge and Imagination and In-image Text and Layout. The distribution of these dimensions within our evaluation dataset is detailed in Figure~\ref{fig:dataset_analyst}, which highlights a significant focus on challenging aspects like artistic style, complex interactions, and full-body actions.

\textbf{Reward Model Training.} The AlignEvaluator is trained on a large-scale dataset of (reprompt, image) pairs, each annotated with scores for these 24 key points. This process yields a robust and explicit reward signal that is essential for effectively training the CoT Rewriter.

%% file: sec/3_0_data_pipe.tex
\begin{figure}[t]
\centering
\includegraphics[width=1.0\linewidth]{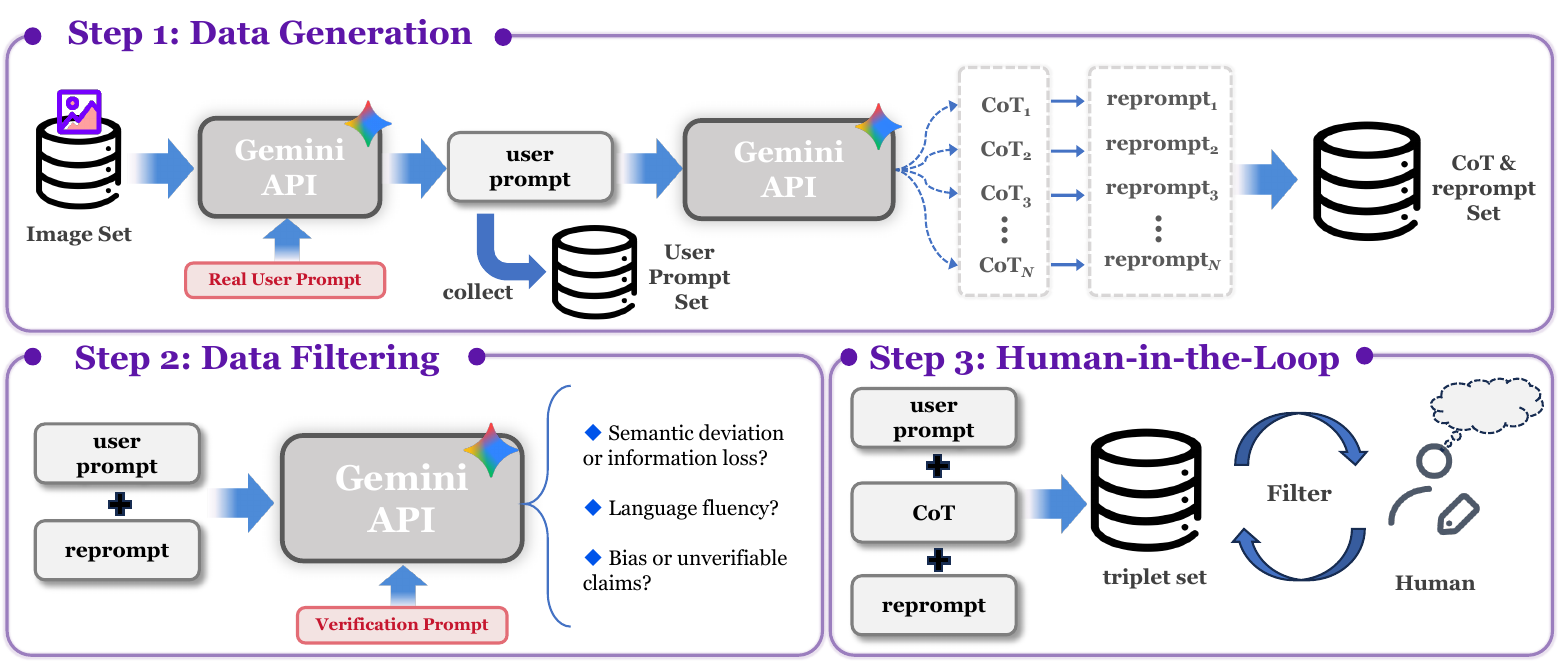}
\caption{
\textbf{Overview of the construction and filtering pipeline for the \methodname training data.}
The process involves user prompt simulation, Gemini-based generation, human-in-the-loop selection, and automated filtering to ensure high quality.
}
\label{fig:data}
\end{figure}

\section{Data Pipeline}
\label{sec:data_pipe}

A high-quality dataset is critical for training an effective prompt rewriter. 
In this section, we present a multi-stage data curation pipeline designed to produce a large-scale, high-fidelity corpus for both Supervised Fine-Tuning (SFT) and subsequent policy alignment. 
The overall process is illustrated in Figure~\ref{fig:data}.

\subsection{SFT Data for Rewriter Initialization}

We construct the SFT dataset through a four-stage process: 
(1) simulating realistic user prompts from a large image pool; 
(2) generating Chain-of-Thought (CoT) and multiple reprompt candidates using a powerful large language model; 
(3) selecting the optimal reprompt via human-in-the-loop evaluation; and 
(4) performing automated quality filtering.
This pipeline yields a final dataset of \textbf{485,119 high-quality instances}, each comprising a (user prompt, CoT, reprompt) triplet.

\begin{figure}[t]
\centering
\includegraphics[width=0.85\linewidth]{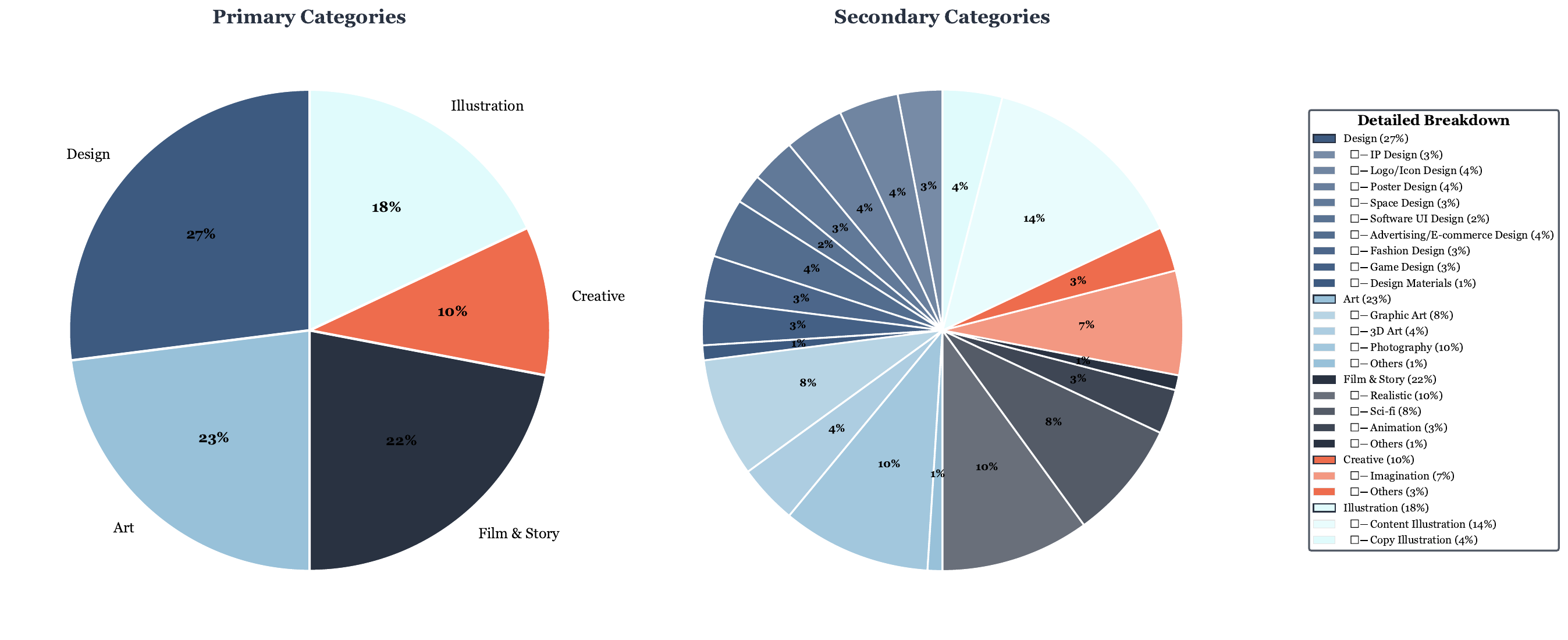}
\vspace{-1em}
\caption{
\textbf{Distribution of Categories in the Dataset.}
The chart on the left shows the primary categories, while the chart on the right provides a detailed breakdown into 20 sub-categories.
}
\label{fig:data_category_distribution}
\end{figure}

Based on fig.~\ref{fig:data_category_distribution}, the dataset exhibits a thematic distribution across several creative domains. Design constitutes the largest portion at 27\%, followed by Art at 23\%, and Film \& Story at 22\%. Illustration accounts for 18\%, and Creative tasks make up 10\% of the dataset.

\textbf{1. User Prompt Simulation.} 
To emulate concise user queries at scale, we begin with a diverse pool of 3.26 million images (1.53M Chinese-centric and 1.73M English-centric). 
We employ an image captioning model to generate short, naturalistic descriptions for these images. 
This step results in a collection of 2.26 million proxy user prompts that serve as the foundation for our data generation.

\textbf{2. CoT and Reprompt Generation.} 
For each simulated user prompt, we leverage the generative power of Gemini-2.5-Pro~\citep{gemini2.5_comanici_2025}. 
Guided by the system prompts detailed in the Appendix, the model produces a detailed CoT reasoning sequence and multiple enriched reprompt candidates. 
This one-to-many generation strategy is crucial for exploring diverse rewriting possibilities, from which we can later select the optimal one.



\textbf{3. Automated Filtering.}
The initial stage involves a rigorous automated filtering process aimed at enhancing data quality.
We utilize Gemini-2.5-Pro to programmatically detect and discard samples with issues such as semantic deviation, information loss, or linguistic incoherence.
This process reduces the dataset from 1 million instances to 611,921 triplets.

\textbf{4. Human-in-the-Loop Selection.}
To further ground our dataset in perceptual quality, we introduce a critical human evaluation step.
For each set of generated reprompt candidates, we use the Hunyuan Text-to-Image model to synthesize corresponding images.
Professional annotators then evaluate these images, selecting the reprompt that best aligns with the user's intent and produces the highest-quality visual output.
This human-in-the-loop process results in a final curated set of \textbf{485,119 triplets}.

\subsection{Prompt Set for Policy Alignment}

For the reinforcement learning stage, we require a distinct set of prompts to drive the GRPO policy optimization. 
We construct a set of approximately 50,000 prompts using the same simulation method as the SFT data but sourced from a disjoint set of images.

\textbf{Data Segregation and Purpose.} 
Crucially, this prompt set is kept entirely separate from the SFT dataset to prevent data leakage and ensure the model learns to generalize its rewriting capabilities. 
Unlike SFT data, these prompts do not have ground-truth CoTs or reprompts; the learning signal is provided dynamically by the AlignEvaluator during online training. 
The thematic distribution of this RL prompt set mirrors that of the SFT data to maintain a consistent optimization landscape.

\begin{figure}[t]
\centering
\includegraphics[width=1.0\linewidth]{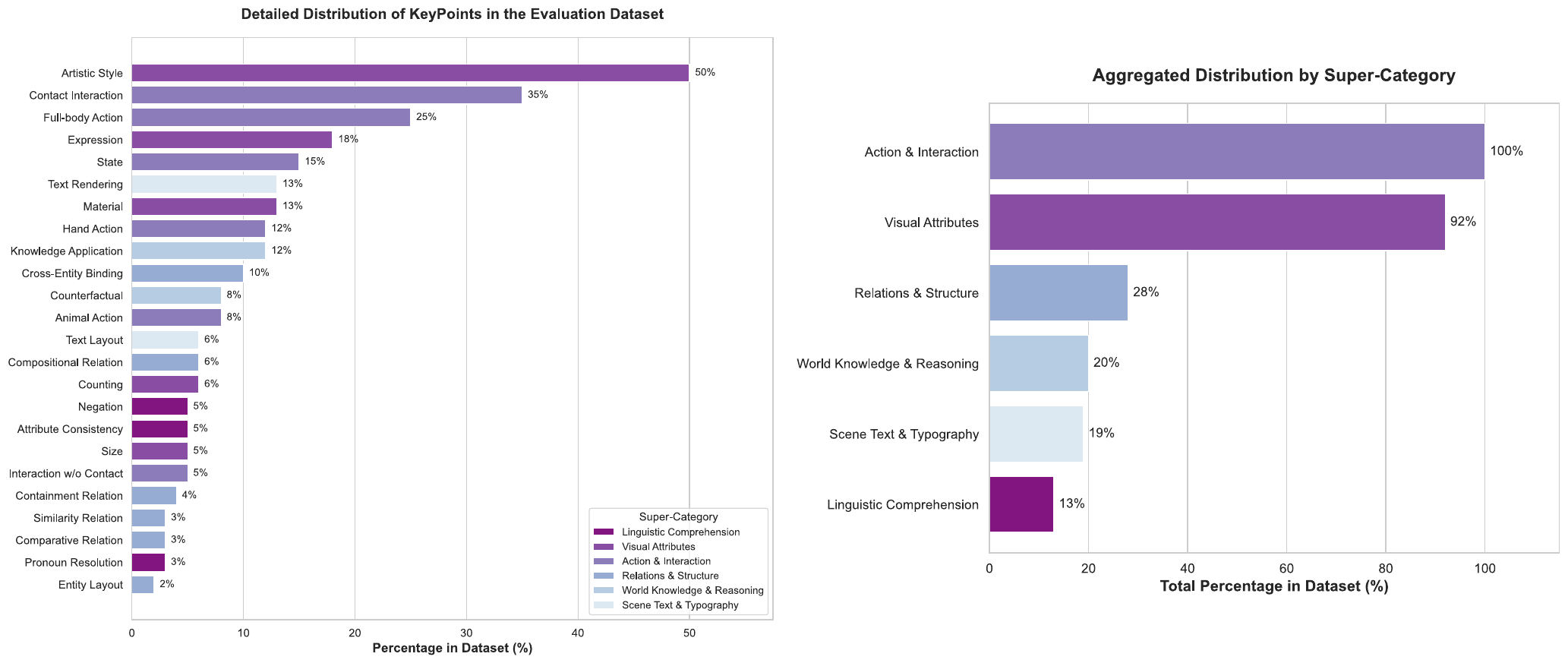}
\caption{
\textbf{Distribution of evaluation dimensions in our dataset.}
(a) The detailed percentage of each of the 24 fine-grained KeyPoints, sorted in descending order. 
(b) The aggregated percentage for each of the six main Super-Categories, calculated by summing the percentages of their constituent KeyPoints. In both charts, colors represent the Super-Category, visually linking the detailed points to their broader classification.
}
\label{fig:dataset_analyst}
\end{figure}
\begin{table}[h]
\caption{A Multi-dimensional Evaluation Framework for Text-to-Image Generation. 
TIC = Text-Image Consistency, SI = Structural Integrity, TIC\&SI = both.}
\label{tab:keypoints}
\centering
\resizebox{0.9\textwidth}{!}{
\small
\renewcommand{\arraystretch}{0.85}
\rowcolors{2}{white}{gray!5}
\begin{tabular}{
  p{0.20\linewidth}  
  p{0.18\linewidth}  
  p{0.40\linewidth}  
  p{0.28\linewidth}  
  p{0.10\linewidth}  
}
\toprule
\rowcolor{gray!15}
\textbf{Super-Category} & 
\textbf{Category} & 
\textbf{Key Point} & 
\textbf{Example} & 
\textbf{Criteria} \\
\midrule

\textbf{Linguistic Comprehension}
& Logical Ops & \textbf{Negation} – interpret negatives 
& Prompt: A bowl of beef noodles, no scallions. (No scallions) 
& TIC \\

& Logical Ops & \textbf{Attribute Consistency} – one attribute bound to many 
& \raggedright Prompt: Five people all wearing red clothes. (All red) 
& TIC \\

& Co-reference & \textbf{Pronoun Resolution} – resolve ambiguity 
& \raggedright Prompt: The large ball broke the table because it was made of metal. (“it” = ball) 
& TIC \\
\midrule

\textbf{Visual Attributes}
& Obj-level & \textbf{Counting} – numeracy ($n \geq 3$) 
& \raggedright Prompt: A picture with four dogs. (Four dogs) 
& TIC \\

& Obj-level & \textbf{Size} – relative comparison 
& \raggedright Prompt: Two large spheres. (Large spheres) 
& TIC \\

& Obj-level & \textbf{Material} – render different materials
& \raggedright Prompt: An ice sculpture of an eagle. (Ice sculpture)
& TIC \\

& Obj-level & \textbf{Expression} – capture facial emotions
& \raggedright Prompt: A strong man, low-angle shot, with a contemptuous expression.
& TIC \\

& Global Style & \textbf{Artistic Style} – adhere to style 
& \raggedright Prompt: Eight galloping horses in Chinese ink wash. 
& TIC \\
\midrule

\textbf{Action \& Interaction}
& \makecell[l]{Individual Action} 
& \makecell[l]{\textbf{Full-body Action} – complex movement} 
& \makecell[l]{Prompt: A girl performing a\\Thomas flare.}
& TIC\&SI \\

& \makecell[l]{Individual Action}
& \makecell[l]{\textbf{Hand Action} – detailed hand/finger structure}
& \raggedright Prompt: A hand using chopsticks to pick up food.
& TIC\&SI \\

& \makecell[l]{Individual Action}
& \makecell[l]{\textbf{Animal Action} – actions performed by animals}
& \raggedright Prompt: A puppy happily running.
& TIC\&SI \\

& Interaction 
& \makecell[l]{\textbf{Contact Interaction} – physical interaction}
& \raggedright Prompt: A boxer lands a punch on a punching bag. 
& TIC\&SI \\

& Interaction
& \makecell[l]{\textbf{Interaction w/o Contact} – non-physical interaction}
& \raggedright Prompt: Einstein looking at Hawking.
& TIC \\

& State
& \makecell[l]{\textbf{State} – continuous state of being or action}
& \raggedright Prompt: A gust of wind blows, cherry blossoms dance in the air.
& TIC \\
\midrule

\textbf{Relations \& Structure}
& Semantic Rel. & \textbf{Comparative Relation} – attribute comparison 
& \raggedright Prompt: Woman in red dress taller than woman in yellow. 
& TIC \\

& Semantic Rel. & \textbf{Compositional Relation} – entity composed of others
& \raggedright Prompt: A cat made of orange slices.
& TIC \\

& Semantic Rel. & \textbf{Containment Relation} – container holds an entity
& \raggedright Prompt: A cup full of soda water.
& TIC \\

& Semantic Rel. & \textbf{Similarity Relation} – resemblance in shape
& \raggedright Prompt: A lake shaped like a guitar.
& TIC \\

& Spatial Layout & \textbf{Cross-Entity Binding} – distinct attributes to entities 
& \raggedright Prompt: Man (buzz cut, blue shirt) and woman (long hair, yellow shirt). 
& TIC \\

& Spatial Layout & \textbf{Entity Layout} – specific arrangement of entities
& \raggedright Prompt: A race car on a city track, with a mini-map in the top-left corner.
& TIC \\
\midrule

\textbf{World Knowledge \& Reasoning}
& \makecell[l]{World\\Knowledge} 
& \makecell[l]{\textbf{Knowledge Application} –\\famous entities}  
& \makecell[l]{Prompt: The Great Wall of\\China / Marie Curie.}
& TIC \\

& \makecell[l]{Abstract\\Reasoning}  & \textbf{Counterfactual} – surreal impossible scenes 
& \raggedright Prompt: A girl held onto the stem of a huge dandelion with both hands, suspended above the clouds.
& TIC \\
\midrule

\textbf{Scene Text \& Typography}
& In-Image Text 
& \makecell[l]{\textbf{Text Rendering} – \\render text content accurately}
& \makecell[l]{Prompt: Poster with text \\ ``Game of Thrones'' at the bottom.}
& TIC \\

& In-Image Text
& \makecell[l]{\textbf{Text Layout} – position text as instructed}
& \raggedright Prompt: Poster of a woman on a throne of waves, text "Game of Thrones" at the bottom.
& TIC \\

\bottomrule
\end{tabular}
}
\end{table}

%% file: sec/3_experiment.tex
\begin{figure}[t]
\centering
\includegraphics[width=0.95\linewidth]{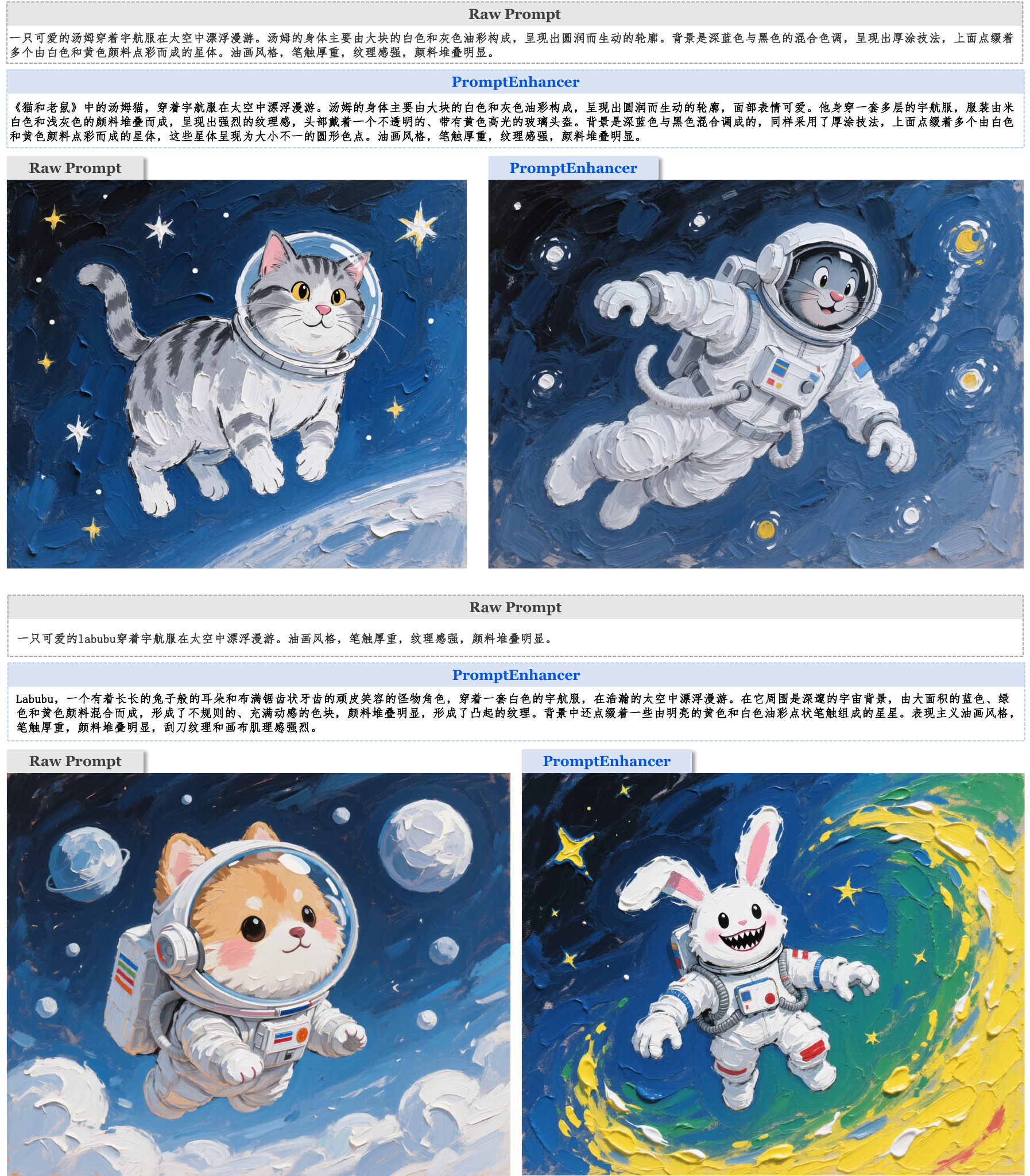}
\caption{
\textbf{
Qualitative Comparison of Prompt Rewriting.
}
This figure demonstrates the effectiveness of the \methodname prompt rewriter. Each comparison pair shows an image generated from a simple "Raw Prompt" alongside an image generated from the detailed prompt created by \methodname. As illustrated, the enriched prompts, which add specific details like character identity (``Tom cat from Tom \& Jerry") and artistic style (``oil painting style, heavy brushstrokes"), guide the model to produce images with significantly greater detail, stylistic accuracy, and fidelity to the user's intent.
}
\label{fig:abl_reprompt_ch}
\end{figure}

\section{T2I-Keypoints Evaluation Benchmark}
\label{sec:benchmark}

To facilitate a comprehensive and fine-grained evaluation of Text-to-Image (T2I) models, we introduce the \textbf{T2I-Keypoints-Align} benchmark.\footnote{The benchmark is publicly available at: \url{https://huggingface.co/datasets/PromptEnhancer/T2I-Keypoints-Eval}} This benchmark is specifically designed to assess a model's ability to accurately render complex prompts by breaking them down into constituent semantic `keypoints`. Our dataset comprises a total of 6,687 prompts, balanced between English (3,000 records, 44.9\%) and Chinese (3,687 records, 55.1\%). Each prompt is meticulously annotated with multiple keypoint categories—such as actions, attributes, and relationships—enabling a granular analysis of model performance across various conceptual dimensions.

\begin{figure}[t]
  \centering
  \begin{subfigure}[t]{0.48\textwidth}
    \centering
    \includegraphics[width=\textwidth]{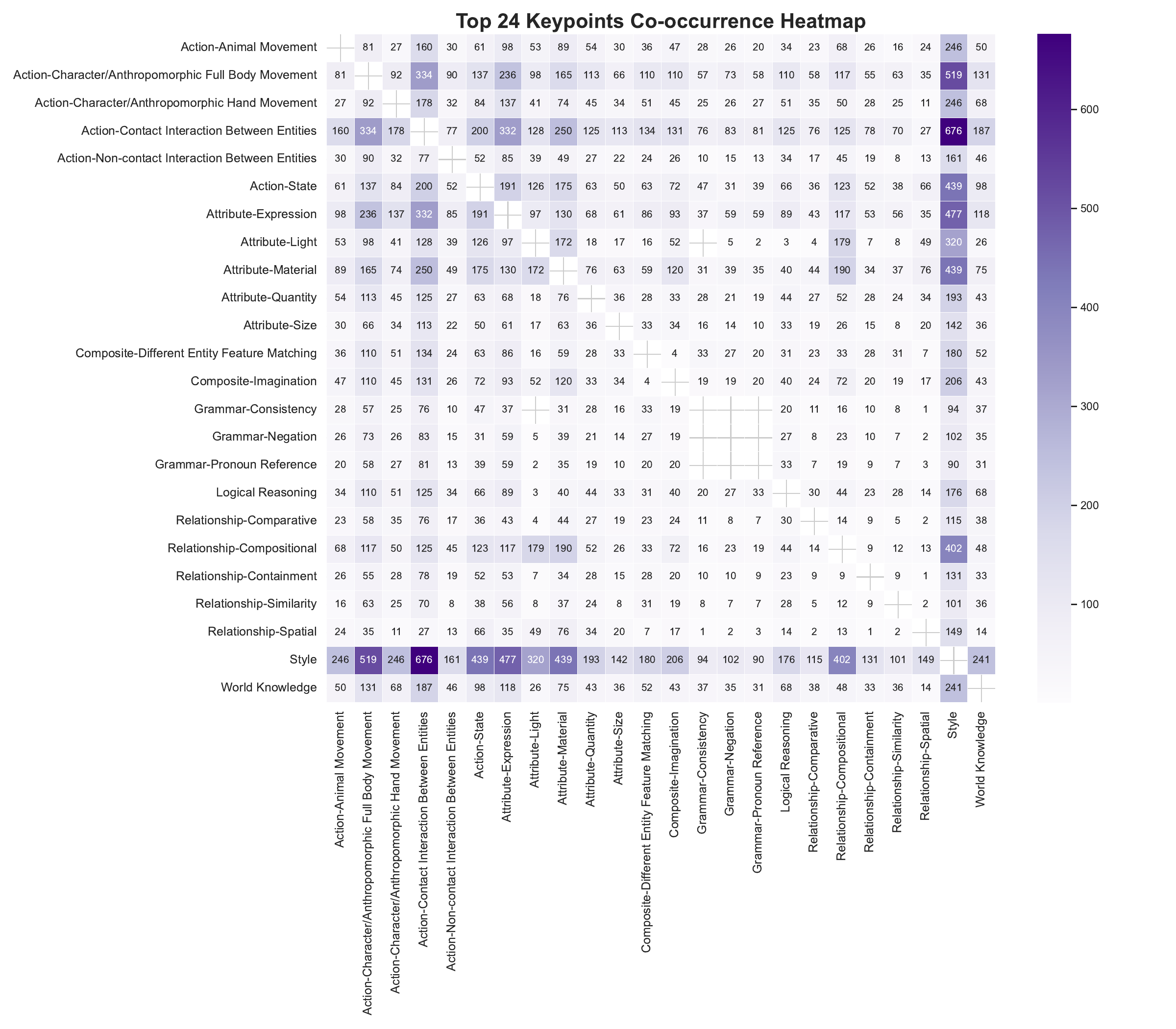}
    \caption{Chinese data}
  \end{subfigure}
  \hfill
  \begin{subfigure}[t]{0.48\textwidth}
    \centering
    \includegraphics[width=\textwidth]{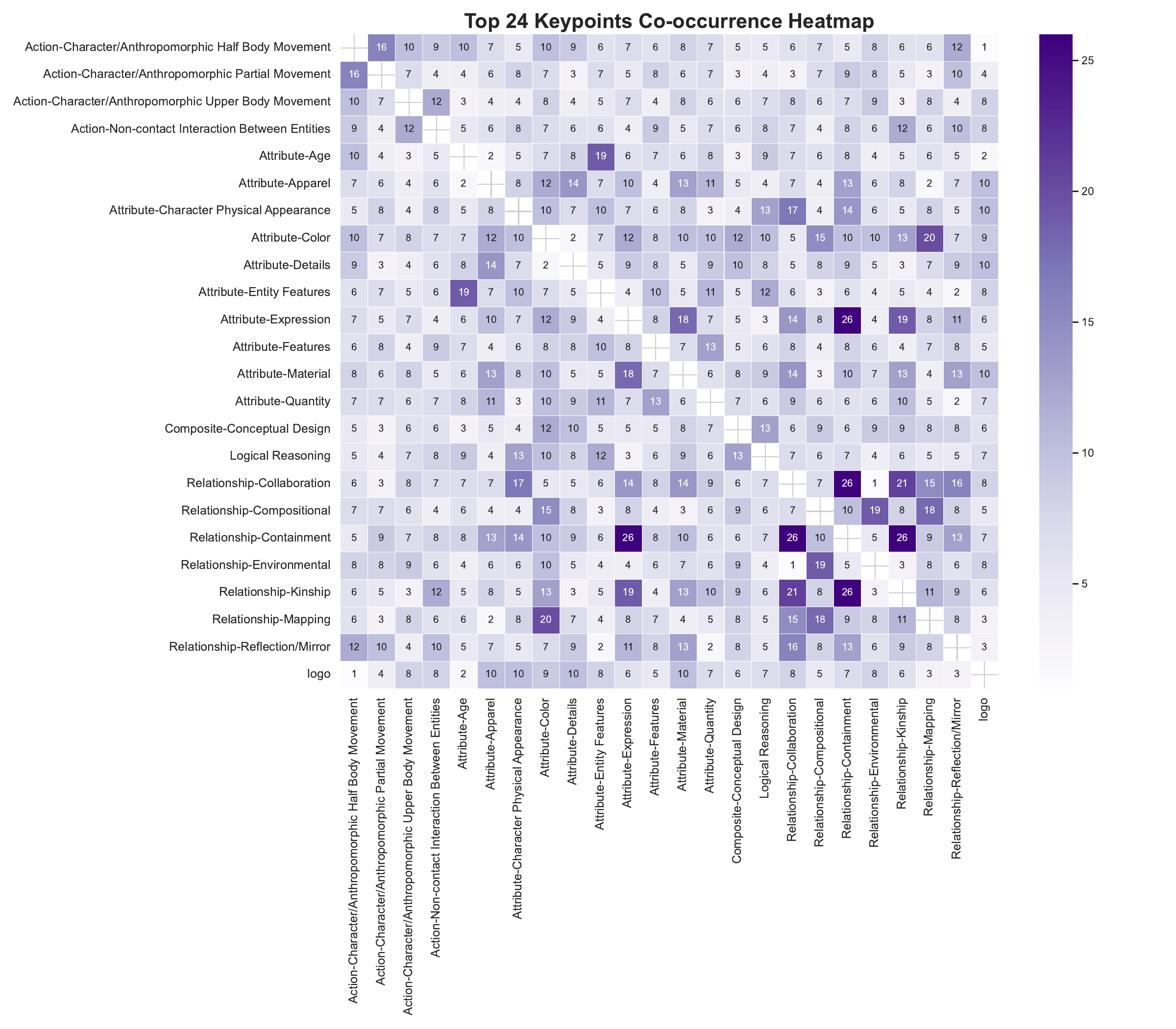}
    \caption{English data}
  \end{subfigure}
  \caption{Co-occurrence heatmaps of the top 24 most frequent keypoints, comparing the Chinese (a) and English (b) portions of our benchmark. The color intensity reflects the co-occurrence frequency. The Chinese data shows strong correlations between `Style` and `World Knowledge`, while the English data displays more varied pairings.}
\label{fig:heatmap}
\end{figure}

\begin{figure}[h]
  \centering
  \begin{subfigure}[t]{0.48\textwidth}
    \centering
    \includegraphics[width=\textwidth]{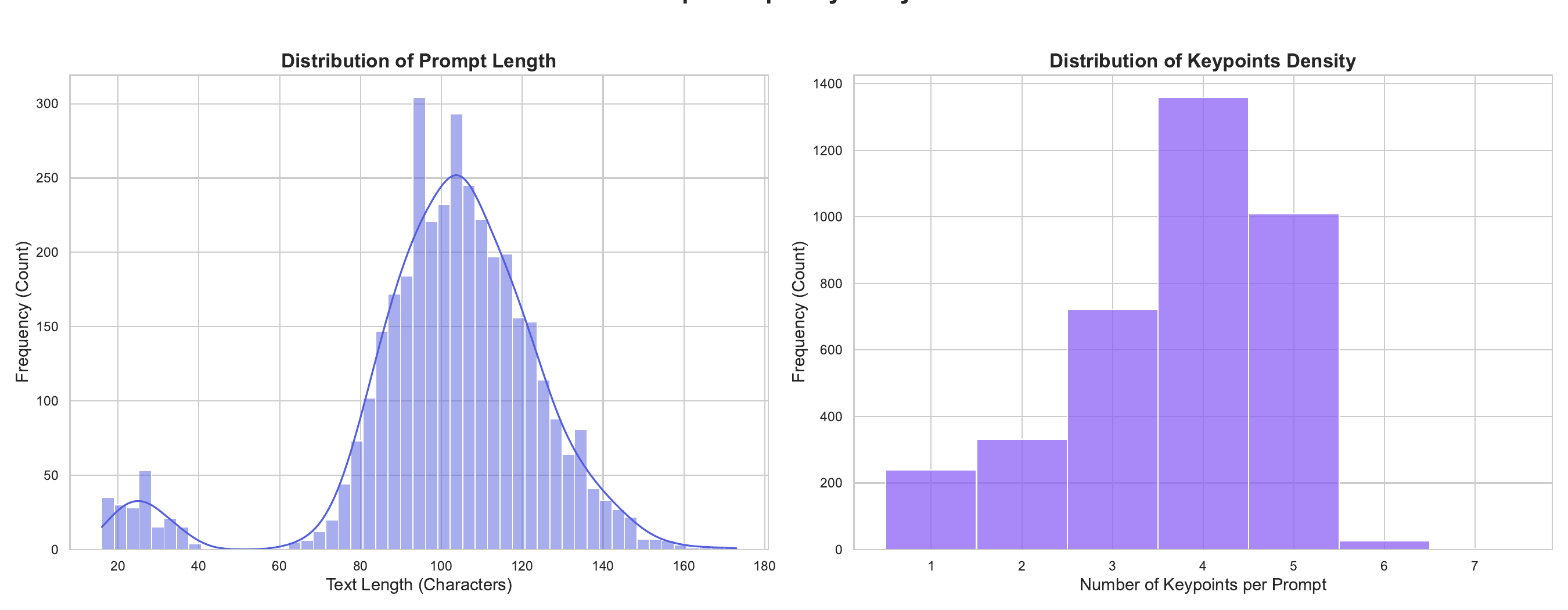}
    \caption{Chinese data}
  \end{subfigure}
  \hfill
  \begin{subfigure}[t]{0.48\textwidth}
    \centering
    \includegraphics[width=\textwidth]{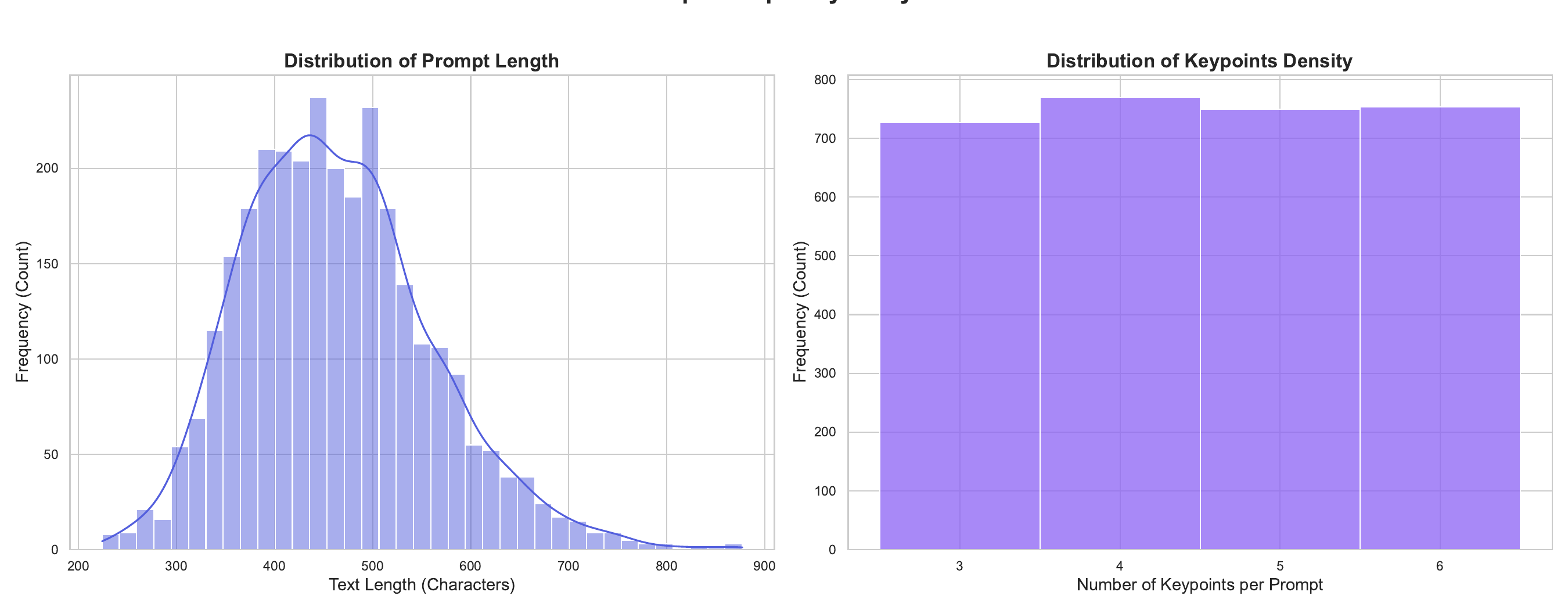}
    \caption{English data}
  \end{subfigure}
  \caption{Statistical distributions of prompt characteristics, comparing the Chinese (a) and English (b) datasets. Each subfigure displays the distribution of prompt length (characters) and keypoint density (number of keypoints per prompt). A significant difference is observed in prompt length, with English prompts being substantially longer and more descriptive.}
\label{fig:prompt_distributions}
\end{figure}

To better understand the characteristics of our benchmark, we conducted a statistical analysis of the prompts and their associated keypoints, with key findings visualized in Figure~\ref{fig:heatmap} and Figure~\ref{fig:prompt_distributions}.


Figure~\ref{fig:prompt_distributions} reveals significant structural differences between the prompts in the two languages. The Chinese prompts (left) are concise, with lengths tightly clustered around a mean of approximately 100 characters. Their keypoint density peaks at 4, with most prompts containing 3 to 5 keypoints. 
Conversely, the English prompts (right) are substantially more verbose and descriptive, with lengths centered around 500 characters. Their keypoint density is more uniformly distributed, with a high and consistent number of prompts containing 3, 4, 5, or 6 keypoints.
This distinction highlights that the English subset tests a model's ability to parse longer narratives with high compositional complexity, while the Chinese subset focuses on efficiently capturing the essence of more succinct requests. Together, they form a robust benchmark for evaluating T2I models against diverse prompt styles.

\section{Experiments}
\label{sec:experiments}
In this section, we conduct a comprehensive set of experiments to validate the efficacy of our proposed \methodnamebf framework. 
Our evaluation is designed to demonstrate that \methodnamebf significantly improves a T2I model's ability to follow complex prompts, and that these improvements are both quantitatively measurable and perceptually preferred by humans.

\subsection{Experimental Setup}
\label{sec:exp_setup}
All experiments are conducted using the \basemodel as the base Text-to-Image model. 
Its weights remain frozen throughout our training process to demonstrate the plug-and-play capability of \methodname. 
Our CoT Rewriter is initialized from the Hunyuan-7B-Instruct model~\citep{HunyuanA13B_tencent_2025}. 
All training and inference are performed on 8 NVIDIA H800 GPUs.


\textbf{Evaluation Benchmark.}
To rigorously assess performance, all methods are evaluated on our newly curated benchmark. This benchmark features challenging prompts specifically designed to test the wide range of fine-grained alignment capabilities detailed in Section~\ref{sec:alignevaluator} and Table~\ref{tab:keypoints}.

\subsection{Implementation Details}
\label{sec:implementation_details}

We detail the training hyperparameters for the two stages of our framework in Table~\ref{tab:hyperparams}.

\textbf{Stage 1: Supervised Fine-Tuning (SFT).}
The CoT Rewriter is fine-tuned on our curated SFT dataset for 10 epochs. We use a learning rate of $1.0 \times 10^{-5}$ with a cosine learning rate scheduler and a warmup ratio of 10\%. The effective batch size is 128, achieved with a per-device batch size of 8 and 2 gradient accumulation steps across 8 GPUs. We leverage bfloat16 mixed-precision training to improve efficiency.

\textbf{Stage 2: Policy Alignment with GRPO.}
Following SFT, the rewriter is further optimized using the GRPO algorithm for 10 epochs. We lower the learning rate to $1.0 \times 10^{-6}$ for stable policy refinement. During the rollout phase for each prompt, we generate $N=8$ candidate reprompts to estimate the policy gradient. A KL-divergence penalty with a coefficient of $0.001$ is applied to regularize the policy updates and prevent deviation from the SFT-initialized model. The global training batch size is 64.

\begin{table}[t]
\centering
\caption{Key hyperparameters for the SFT and GRPO training stages.}
\label{tab:hyperparams}
\resizebox{0.6\textwidth}{!}{%
\begin{tabular}{lcc}
\toprule
\textbf{Hyperparameter} & \textbf{SFT Stage} & \textbf{GRPO Stage} \\
\midrule
Base Model & Hunyuan-7B-Instruct & SFT-tuned Rewriter \\
Learning Rate & $1.0 \times 10^{-5}$ & $1.0 \times 10^{-6}$ \\
LR Scheduler & Cosine & Constant \\
Warmup Ratio & 0.1 & N/A \\
Epochs & 2 & 1 \\
Effective Batch Size & 128 & 64 \\
Precision & bfloat16 & bfloat16 \\
Rollout Samples ($N$) & N/A & 8 \\
KL Coefficient & N/A & 0.001 \\
\bottomrule
\end{tabular}%
}
\end{table}

\subsection{Quantitative Analysis}
\label{sec:quantitative}

\begin{figure}[h]
\centering
\includegraphics[width=1.0\linewidth]{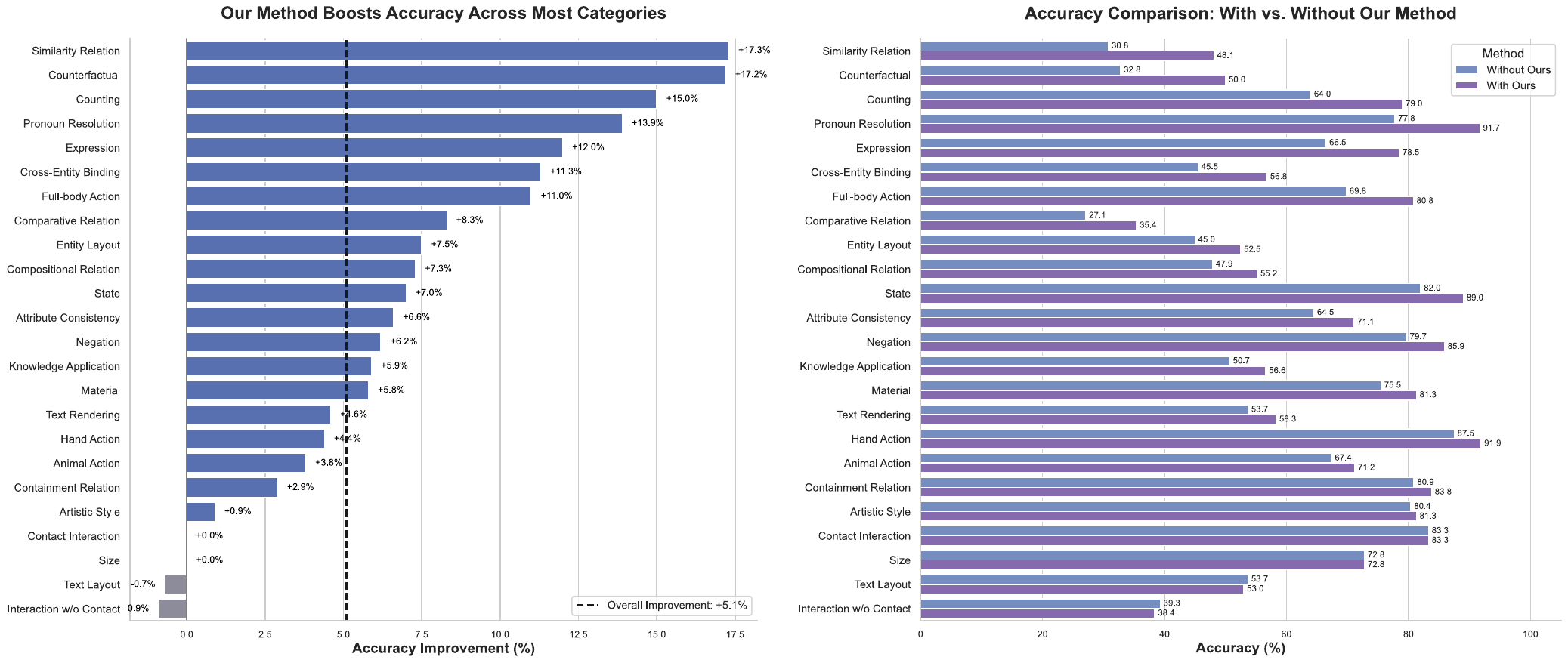}
\caption{
\textbf{Quantitative Evaluation of \methodname's Impact on Prompt Following Accuracy.}
The figure presents a comparative analysis of text-to-image generation accuracy with and without the \methodname framework across 24 distinct semantic categories. The left panel illustrates the percentage point (pp) improvement for each category, highlighting significant gains (blue) in areas like grammatical understanding and compositional reasoning, as well as regressions (red) in others. The right panel provides a direct comparison of the absolute accuracy scores, showing the performance of the baseline model (``w/o Ours") versus the enhanced model (``w/ Our").
}
\label{fig:abl_acc}
\end{figure}

As illustrated in Figure~\ref{fig:abl_acc}, our framework provides a significant and broad-spectrum enhancement to the T2I model's prompt-following capabilities. 
On average, \methodnamebf boosts the accuracy across all 24 evaluation dimensions by a substantial \textbf{+5.1\%}. 
This overall improvement is underpinned by widespread gains, with performance increasing in 20 out of 24 categories.

\textbf{Significant Gains in Complex Reasoning.}
The most pronounced improvements are concentrated in categories that demand sophisticated semantic and compositional understanding. 
Notably, we observe dramatic boosts in \textbf{Similarity Relation} (+17.3\%), \textbf{Counterfactual} reasoning (+17.2\%), and \textbf{Counting} (+15.0\%). 
These results underscore our method's ability to help the T2I model interpret abstract relationships and precise numerical constraints. 
Furthermore, significant enhancements are seen in fine-grained linguistic and visual challenges, including \textbf{Pronoun Resolution} (+13.9\%), rendering nuanced facial \textbf{Expression} (+12.0\%), and maintaining \textbf{Cross-Entity Binding} (+11.3\%). 
In total, 15 distinct categories exhibit an accuracy improvement of more than 5.0\%, demonstrating the comprehensive impact of our prompt rewriting strategy.

\textbf{Areas of Neutrality and Minor Regression.}
While the impact is overwhelmingly positive, the gains are not uniform. 
Performance in \textbf{Contact Interaction} and \textbf{Size} remains unchanged (+0.0\%), and \textbf{Artistic Style} shows only a marginal improvement (+0.9\%), suggesting the baseline model is already relatively competent in these areas. 
We identify minor performance regressions in two categories: \textbf{Text Layout} (-0.7\%) and \textbf{Interaction w/o Contact} (-0.9\%). 
These isolated cases represent valuable areas for future work, indicating that for certain simple concepts, the rewriting process might occasionally introduce ambiguity or over-specification. 
Nevertheless, these minor trade-offs are far outweighed by the extensive and significant improvements across the vast majority of challenging categories.

\begin{figure}[t]
\centering
\includegraphics[width=0.9\linewidth]{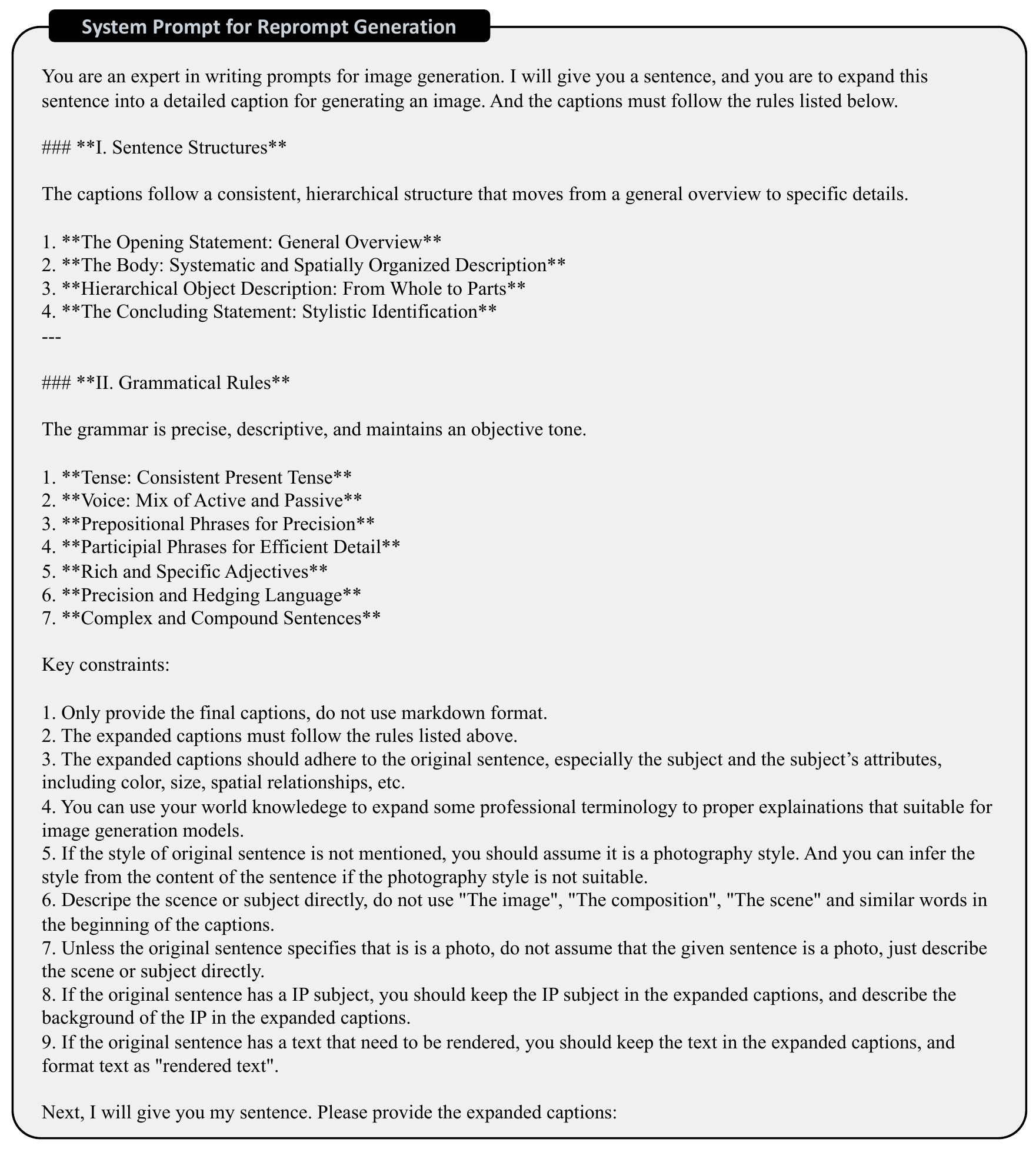}
\caption{
The system prompt designed to guide Gemini-2.5-Pro for ``Reprompt Generation''. This prompt expands a simple input sentence into a structured, detailed description suitable for image generation, following a sophisticated framework. 
This framework consists of three core parts: \textbf{I. Sentence Structures} (defining a four-level descriptive hierarchy from macro to micro), \textbf{II. Grammatical Rules} (specifying seven conventions to ensure objectivity and precision), and \textbf{9 key constraints}, which together guarantee the quality and consistency of the generated content.}
\label{fig:template_reprompt}
\end{figure}

\begin{figure}[t]
\centering
\includegraphics[width=0.9\linewidth]{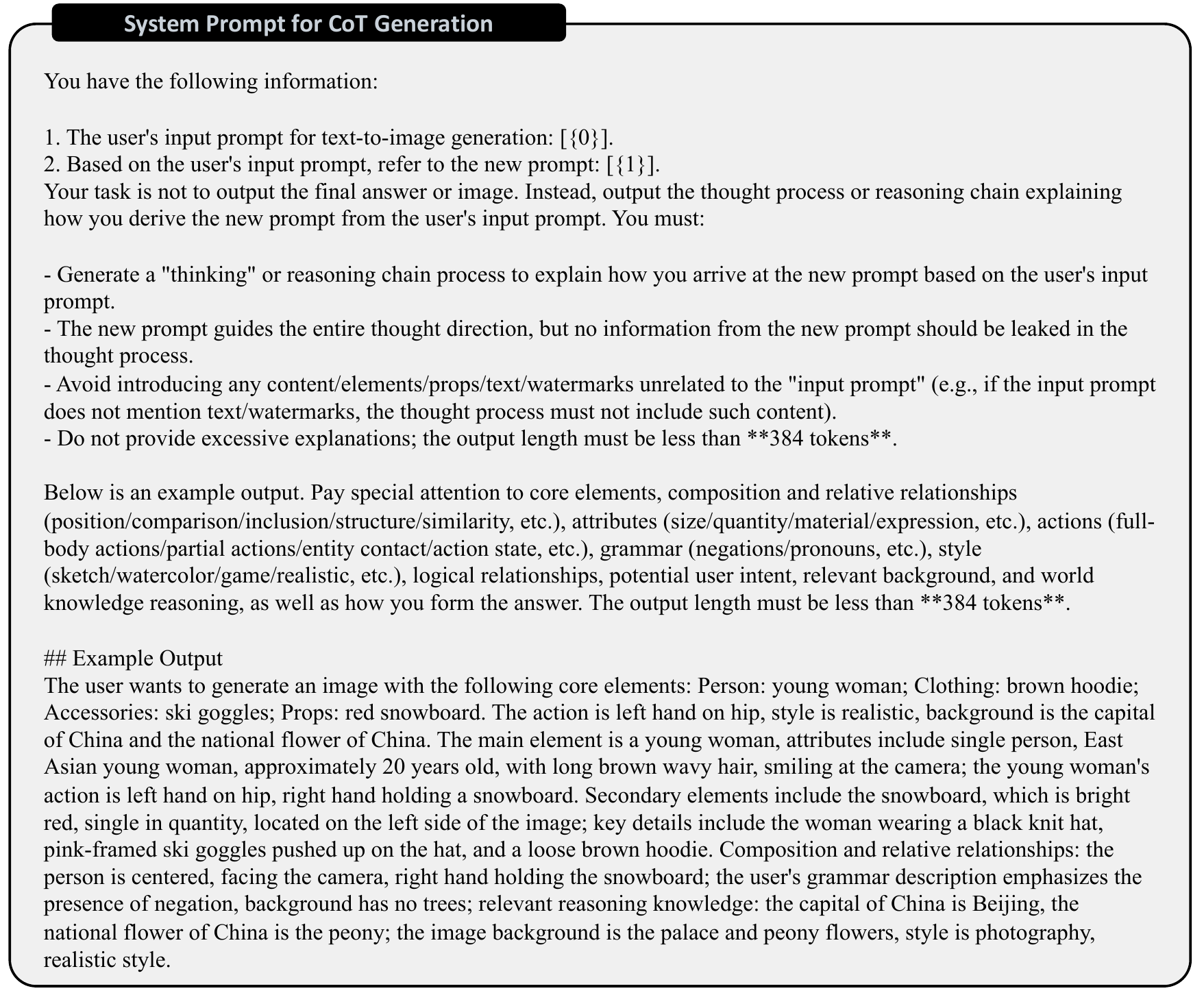}
\caption{
The system prompt for generating a Chain-of-Thought. 
The core task is not to output a final answer, but to generate a reasoning process that explains how the system derives a new, optimized prompt (\texttt{[\{1\}]}) from the user's initial input (\texttt{[\{0\}]}). The prompt requires the model to focus on a series of analytical dimensions (\eg, core elements, composition, attributes, style, world knowledge) and specifies the required format and depth through a detailed ``Example Output''.
}
\label{fig:template_cot_prompt}
\end{figure}

%% file: sec/4_relatedwork.tex
\section{Related Work}

\subsection{Text-to-Image Generation}
The field of Text-to-Image (T2I) synthesis has witnessed a paradigm shift with the advent of diffusion models~\citep{ddpm_ho_2020,song_scorebased_2021}, which have largely superseded earlier GAN-based approaches~\citep{generative_reed_2016}. 
Seminal models like DALL-E~\citep{DALLE_ramesh_2021}, Imagen~\citep{imagen_saharia_2022}, and Stable Diffusion~\citep{LDM_rombach_2022,sdxl_podell_2024,sd35_esser_2024} established a powerful foundation for generating high-fidelity and diverse images. This frontier continues to be pushed by state-of-the-art systems such as DALL-E 3~\citep{betker2023improving}, and more recent architectures like FLUX.1~\citep{Fluxkontext_blacklabs_2025}, which have dramatically improved photorealism and text rendering.

However, despite this progress, a fundamental challenge persists: the prompt alignment gap. Models often struggle with compositional complexity, failing to correctly bind attributes to objects, comprehend spatial or logical relationships (\eg, negation), and maintain fidelity to intricate details~\citep{t2icompbench_huang_2023}. 
This gap places a significant ``alignment tax'' on users, forcing them to engage in tedious and often unintuitive prompt engineering. Our work directly targets this gap, not by inventing a new generator, but by creating a universal interface that teaches any T2I model to better understand human intent.

\subsection{Prompt Rewriting and Optimization}
To bridge the user-model communication gap, researchers have increasingly focused on prompt optimization. One foundational strategy is \textit{recaptioning}, where high-quality, descriptive captions are synthesized to fine-tune the T2I model itself, as famously implemented in DALL-E 3~\citep{betker2023improving}.

A more direct approach is on-the-fly prompt rewriting, where user prompts are enhanced before being passed to a frozen T2I model. These methods include: 
(1) \textit{Iterative Refinement}, where an LLM critiques and revises a prompt in a loop to improve the generated output~\citep{wu2024self,yang2024idea2img}; 
(2) \textit{Automatic Prompt Engineering}, which learns to generate or select stylistically potent or compositionally robust prompts~\citep{cao2023beautifulprompt,manas2024improving}; and 
(3) \textit{Agent-based Systems}, which employ multimodal LLM agents to plan, generate, and edit in complex workflows~\citep{wang2024genartist,qin2024diffusiongpt}. 

While powerful, these methods often rely on general-purpose LLMs (\eg, GPT-4~\citep{gpt4o_hurst_2024}) that lack a specialized understanding of T2I-specific failure modes. Our approach charts a new path by training a dedicated rewriter that is explicitly optimized against a fine-grained, T2I-centric reward model, enabling it to learn the nuances of visual generation.

\subsection{Chain-of-Thought for Controllable Generation}
Chain-of-Thought (CoT) reasoning~\citep{CoT_wei_2022}, which decomposes complex problems into intermediate steps, has proven highly effective for improving controllability in multimodal generation. Instead of asking a model to synthesize a complex scene in one shot, CoT-inspired methods impose a structured, step-wise process. For instance, some frameworks leverage a powerful multimodal LLM to first generate an explicit \textit{plan}—such as a scene layout, object coordinates, or a sequence of rendering steps—which then guides a constrained diffusion process to enhance compositional accuracy~\citep{yang2024mastering, zhang2025layercraft}. Other approaches integrate step-wise reasoning directly into the generative model's architecture~\citep{wang2025mint}.

These methods have significantly advanced compositional fidelity. However, they often create a tight coupling between the reasoner and a specific T2I model architecture. Our framework innovates by leveraging CoT purely within the \textit{prompt rewriting process}. We create a decoupled CoT rewriter that is model-agnostic and trained via reinforcement learning. This allows it to be universally applied as a plug-and-play module to enhance existing, pre-trained T2I models without any architectural modifications.

\subsection{Fine-Grained Evaluation and Reward Modeling}
The maturation of T2I models has necessitated a parallel evolution in evaluation methodologies, moving beyond holistic metrics like FID or CLIP Score. A major advancement was the development of reward models trained on large-scale human preferences, such as ImageReward~\citep{imagereward_xu_2023}, HPSv2~\citep{HPS_Wu_2023,HPSv2_Wu_2023}, which capture a more human-aligned sense of aesthetic quality and overall prompt fidelity.

However, a single preference score can obscure critical, fine-grained errors in compositionality. This has spurred the creation of specialized benchmarks with detailed, multi-faceted annotations, such as T2I-CompBench~\citep{t2icompbench_huang_2023} and EvalMuse~\citep{Evalmuse40k_han_2024}, which probe specific capabilities like attribute binding and spatial relations.

Our work makes a key contribution by closing the loop between fine-grained evaluation and model optimization. We first introduce the `AlignEvaluator`, a reward model trained on a new, comprehensive taxonomy of 24 key points that synthesizes and expands upon existing criteria. Crucially, we then use this detailed, multi-faceted reward signal to directly optimize our prompt rewriter via reinforcement learning. This transforms fine-grained evaluation from a passive measurement tool into an active training signal, directly teaching the rewriter to avoid common T2I failure modes.

%% file: sec/5_conclusion.tex
\section{Conclusion}
In this paper, we introduced \methodname, a novel framework to help text-to-image (T2I) models better understand complex user prompts. T2I models often struggle to follow detailed instructions, leading to images that don't match the user's intent. 
Our method fixes this by automatically rewriting the user's initial prompt into a more detailed one that any T2I model can easily interpret.
Our key innovation is a prompt rewriter that uses a Chain-of-Thought (CoT) process. 
We train this rewriter using reinforcement learning, guided by a custom reward model we call the AlignEvaluator. 
This evaluator provides specific, fine-grained feedback on 24 different aspects of image-text alignment, allowing the rewriter to learn how to create high-quality prompts. Crucially, our framework is universal and works with any pre-trained T2I model without needing to modify it.
Experiments show that \methodname significantly improves the alignment between the generated image and the user's prompt across a wide range of challenges. By decoupling the task of prompt enhancement from image generation, our work offers an effective and scalable solution to improve the control and accuracy of T2I systems.

%% file: sec/x_contributor.tex
\section{Contributors and Acknowledgements}
\begin{itemize}
    \item \textbf{Project leaders:} Qinglin Lu\textsuperscript{‡}, Chunyu Wang\textsuperscript{†}
    \item \textbf{Core Contributors:} Linqing Wang\textsuperscript{*}, Ximing Xing\textsuperscript{*}
    \item \textbf{Contributors:} Yiji Cheng, Zhiyuan Zhao, Donghao Li, Tiankai Hang, Jiale Tao, Qixun Wang, Ruihuang Li, Comi Chen, Xin Li, Mingrui Wu, Xinchi Deng, Shuyang Gu
    \item \textbf{Acknowledgements:} We would like to thank Kai Song, Zhengkai Jiang for their valuable inputs and suggestions.
\end{itemize}

\bigskip
\textsuperscript{*}\,\textbf{Equal Contribution.} \\
\noindent\textsuperscript{†}\,\textbf{Project Lead}: Chunyu Wang \\
\textsuperscript{‡}\,\textbf{Corresponding Author}: Qinglin Lu

%% file: main.bbl
\begin{thebibliography}{37}
\providecommand{\natexlab}[1]{#1}
\providecommand{\url}[1]{\texttt{#1}}
\expandafter\ifx\csname urlstyle\endcsname\relax
  \providecommand{\doi}[1]{doi: #1}\else
  \providecommand{\doi}{doi: \begingroup \urlstyle{rm}\Url}\fi

\bibitem[Betker et~al.(2023)Betker, Goh, Jing, Brooks, Wang, Li, Ouyang, Zhuang, Lee, Guo, et~al.]{betker2023improving}
James Betker, Gabriel Goh, Li~Jing, Tim Brooks, Jianfeng Wang, Linjie Li, Long Ouyang, Juntang Zhuang, Joyce Lee, Yufei Guo, et~al.
\newblock Improving image generation with better captions.
\newblock \emph{Computer Science. https://cdn. openai. com/papers/dall-e-3. pdf}, 2\penalty0 (3):\penalty0 8, 2023.

\bibitem[Cao et~al.(2023)Cao, Wang, Liu, Wu, Zhu, and Huang]{cao2023beautifulprompt}
Tingfeng Cao, Chengyu Wang, Bingyan Liu, Ziheng Wu, Jinhui Zhu, and Jun Huang.
\newblock Beautifulprompt: Towards automatic prompt engineering for text-to-image synthesis.
\newblock \emph{arXiv preprint arXiv:2311.06752}, 2023.

\bibitem[Comanici et~al.(2025)Comanici, Bieber, Schaekermann, Pasupat, Sachdeva, Dhillon, Blistein, Ram, Zhang, Rosen, et~al.]{gemini2.5_comanici_2025}
Gheorghe Comanici, Eric Bieber, Mike Schaekermann, Ice Pasupat, Noveen Sachdeva, Inderjit Dhillon, Marcel Blistein, Ori Ram, Dan Zhang, Evan Rosen, et~al.
\newblock Gemini 2.5: Pushing the frontier with advanced reasoning, multimodality, long context, and next generation agentic capabilities.
\newblock \emph{arXiv preprint arXiv:2507.06261}, 2025.

\bibitem[Esser et~al.(2024)Esser, Kulal, Blattmann, Entezari, M{\"u}ller, Saini, Levi, Lorenz, Sauer, Boesel, et~al.]{sd35_esser_2024}
Patrick Esser, Sumith Kulal, Andreas Blattmann, Rahim Entezari, Jonas M{\"u}ller, Harry Saini, Yam Levi, Dominik Lorenz, Axel Sauer, Frederic Boesel, et~al.
\newblock Scaling rectified flow transformers for high-resolution image synthesis.
\newblock In \emph{Forty-first international conference on machine learning (ICML)}, 2024.

\bibitem[Han et~al.(2024)Han, Fan, Fu, Li, Li, Cui, Wang, Tai, Sun, Guo, et~al.]{Evalmuse40k_han_2024}
Shuhao Han, Haotian Fan, Jiachen Fu, Liang Li, Tao Li, Junhui Cui, Yunqiu Wang, Yang Tai, Jingwei Sun, Chunle Guo, et~al.
\newblock Evalmuse-40k: A reliable and fine-grained benchmark with comprehensive human annotations for text-to-image generation model evaluation.
\newblock \emph{arXiv preprint arXiv:2412.18150}, 2024.

\bibitem[Hao et~al.(2023)Hao, Chi, Dong, and Wei]{hao2023optimizing}
Yaru Hao, Zewen Chi, Li~Dong, and Furu Wei.
\newblock Optimizing prompts for text-to-image generation.
\newblock \emph{Advances in Neural Information Processing Systems}, 36:\penalty0 66923--66939, 2023.

\bibitem[Ho et~al.(2020)Ho, Jain, and Abbeel]{ddpm_ho_2020}
Jonathan Ho, Ajay Jain, and Pieter Abbeel.
\newblock Denoising diffusion probabilistic models.
\newblock \emph{Advances in neural information processing systems (NeurIPS)}, 33:\penalty0 6840--6851, 2020.

\bibitem[Huang et~al.(2023)Huang, Sun, Xie, Li, and Liu]{t2icompbench_huang_2023}
Kaiyi Huang, Kaiyue Sun, Enze Xie, Zhenguo Li, and Xihui Liu.
\newblock T2i-compbench: A comprehensive benchmark for open-world compositional text-to-image generation.
\newblock \emph{Advances in Neural Information Processing Systems (NeurIPS)}, 36:\penalty0 78723--78747, 2023.

\bibitem[Hurst et~al.(2024)Hurst, Lerer, Goucher, Perelman, Ramesh, Clark, Ostrow, Welihinda, Hayes, Radford, et~al.]{gpt4o_hurst_2024}
Aaron Hurst, Adam Lerer, Adam~P Goucher, Adam Perelman, Aditya Ramesh, Aidan Clark, AJ~Ostrow, Akila Welihinda, Alan Hayes, Alec Radford, et~al.
\newblock Gpt-4o system card.
\newblock \emph{arXiv preprint arXiv:2410.21276}, 2024.

\bibitem[Labs(2024)]{Flux_blacklabs_2025}
Black~Forest Labs.
\newblock Flux, 2024.
\newblock URL \url{https://github.com/black-forest-labs/flux}.

\bibitem[Labs et~al.(2025)Labs, Batifol, Blattmann, Boesel, Consul, Diagne, Dockhorn, English, English, Esser, et~al.]{Fluxkontext_blacklabs_2025}
Black~Forest Labs, Stephen Batifol, Andreas Blattmann, Frederic Boesel, Saksham Consul, Cyril Diagne, Tim Dockhorn, Jack English, Zion English, Patrick Esser, et~al.
\newblock Flux. 1 kontext: Flow matching for in-context image generation and editing in latent space.
\newblock \emph{arXiv preprint arXiv:2506.15742}, 2025.

\bibitem[Li et~al.(2024)Li, Zhang, Lin, Xiong, Long, Deng, Zhang, Liu, Huang, Xiao, et~al.]{hunyuandit_li_2024}
Zhimin Li, Jianwei Zhang, Qin Lin, Jiangfeng Xiong, Yanxin Long, Xinchi Deng, Yingfang Zhang, Xingchao Liu, Minbin Huang, Zedong Xiao, et~al.
\newblock Hunyuan-dit: A powerful multi-resolution diffusion transformer with fine-grained chinese understanding.
\newblock \emph{arXiv preprint arXiv:2405.08748}, 2024.

\bibitem[Liu et~al.(2024)Liu, Feng, Xue, Wang, Wu, Lu, Zhao, Deng, Zhang, Ruan, et~al.]{deepseekv3_liu_2024}
Aixin Liu, Bei Feng, Bing Xue, Bingxuan Wang, Bochao Wu, Chengda Lu, Chenggang Zhao, Chengqi Deng, Chenyu Zhang, Chong Ruan, et~al.
\newblock Deepseek-v3 technical report.
\newblock \emph{arXiv preprint arXiv:2412.19437}, 2024.

\bibitem[Ma{\~n}as et~al.(2024)Ma{\~n}as, Astolfi, Hall, Ross, Urbanek, Williams, Agrawal, Romero-Soriano, and Drozdzal]{manas2024improving}
Oscar Ma{\~n}as, Pietro Astolfi, Melissa Hall, Candace Ross, Jack Urbanek, Adina Williams, Aishwarya Agrawal, Adriana Romero-Soriano, and Michal Drozdzal.
\newblock Improving text-to-image consistency via automatic prompt optimization.
\newblock \emph{arXiv preprint arXiv:2403.17804}, 2024.

\bibitem[Peebles \& Xie(2023)Peebles and Xie]{dit_peebles_2023}
William Peebles and Saining Xie.
\newblock Scalable diffusion models with transformers.
\newblock In \emph{Proceedings of the IEEE/CVF international conference on computer vision (ICCV)}, pp.\  4195--4205, 2023.

\bibitem[Podell et~al.(2024)Podell, English, Lacey, Blattmann, Dockhorn, M{\"u}ller, Penna, and Rombach]{sdxl_podell_2024}
Dustin Podell, Zion English, Kyle Lacey, Andreas Blattmann, Tim Dockhorn, Jonas M{\"u}ller, Joe Penna, and Robin Rombach.
\newblock {SDXL}: Improving latent diffusion models for high-resolution image synthesis.
\newblock In \emph{The Twelfth International Conference on Learning Representations (ICLR)}, 2024.
\newblock URL \url{https://openreview.net/forum?id=di52zR8xgf}.

\bibitem[Qin et~al.(2024)Qin, Wu, Chen, Ren, Li, Wu, Xiao, Wang, and Wen]{qin2024diffusiongpt}
Jie Qin, Jie Wu, Weifeng Chen, Yuxi Ren, Huixia Li, Hefeng Wu, Xuefeng Xiao, Rui Wang, and Shilei Wen.
\newblock Diffusiongpt: Llm-driven text-to-image generation system.
\newblock \emph{arXiv preprint arXiv:2401.10061}, 2024.

\bibitem[Radford et~al.(2021)Radford, Kim, Hallacy, Ramesh, Goh, Agarwal, Sastry, Askell, Mishkin, Clark, et~al.]{CLIP_radford_2021}
Alec Radford, Jong~Wook Kim, Chris Hallacy, Aditya Ramesh, Gabriel Goh, Sandhini Agarwal, Girish Sastry, Amanda Askell, Pamela Mishkin, Jack Clark, et~al.
\newblock Learning transferable visual models from natural language supervision.
\newblock In \emph{International conference on machine learning (ICML)}, pp.\  8748--8763. PmLR, 2021.

\bibitem[Ramesh et~al.(2021)Ramesh, Pavlov, Goh, Gray, Voss, Radford, Chen, and Sutskever]{DALLE_ramesh_2021}
Aditya Ramesh, Mikhail Pavlov, Gabriel Goh, Scott Gray, Chelsea Voss, Alec Radford, Mark Chen, and Ilya Sutskever.
\newblock Zero-shot text-to-image generation.
\newblock In \emph{International conference on machine learning (ICML)}, pp.\  8821--8831. Pmlr, 2021.

\bibitem[Reed et~al.(2016)Reed, Akata, Yan, Logeswaran, Schiele, and Lee]{generative_reed_2016}
Scott Reed, Zeynep Akata, Xinchen Yan, Lajanugen Logeswaran, Bernt Schiele, and Honglak Lee.
\newblock Generative adversarial text to image synthesis.
\newblock In \emph{International conference on machine learning (ICML)}, pp.\  1060--1069. Pmlr, 2016.

\bibitem[Rombach et~al.(2022)Rombach, Blattmann, Lorenz, Esser, and Ommer]{LDM_rombach_2022}
Robin Rombach, Andreas Blattmann, Dominik Lorenz, Patrick Esser, and Bj{\"o}rn Ommer.
\newblock High-resolution image synthesis with latent diffusion models.
\newblock In \emph{Proceedings of the IEEE/CVF conference on computer vision and pattern recognition (CVPR)}, pp.\  10684--10695, 2022.

\bibitem[Saharia et~al.(2022)Saharia, Chan, Saxena, Li, Whang, Denton, Ghasemipour, Gontijo~Lopes, Karagol~Ayan, Salimans, et~al.]{imagen_saharia_2022}
Chitwan Saharia, William Chan, Saurabh Saxena, Lala Li, Jay Whang, Emily~L Denton, Kamyar Ghasemipour, Raphael Gontijo~Lopes, Burcu Karagol~Ayan, Tim Salimans, et~al.
\newblock Photorealistic text-to-image diffusion models with deep language understanding.
\newblock \emph{Advances in neural information processing systems (NeurIPS)}, 35:\penalty0 36479--36494, 2022.

\bibitem[Shao et~al.(2024)Shao, Wang, Zhu, Xu, Song, Bi, Zhang, Zhang, Li, Wu, et~al.]{deepseekmath_shao_2024}
Zhihong Shao, Peiyi Wang, Qihao Zhu, Runxin Xu, Junxiao Song, Xiao Bi, Haowei Zhang, Mingchuan Zhang, YK~Li, Yang Wu, et~al.
\newblock Deepseekmath: Pushing the limits of mathematical reasoning in open language models.
\newblock \emph{arXiv preprint arXiv:2402.03300}, 2024.

\bibitem[Song et~al.(2021)Song, Sohl-Dickstein, Kingma, Kumar, Ermon, and Poole]{song_scorebased_2021}
Yang Song, Jascha Sohl-Dickstein, Diederik~P Kingma, Abhishek Kumar, Stefano Ermon, and Ben Poole.
\newblock Score-based generative modeling through stochastic differential equations.
\newblock In \emph{International Conference on Learning Representations (ICLR)}, 2021.
\newblock URL \url{https://openreview.net/forum?id=PxTIG12RRHS}.

\bibitem[Tencent-Hunyuan-Team(2025)]{HunyuanA13B_tencent_2025}
Tencent-Hunyuan-Team.
\newblock Hunyuan-a13b, 2025.
\newblock URL \url{https://github.com/Tencent-Hunyuan/Hunyuan-A13B}.

\bibitem[Wang et~al.(2025)Wang, Liu, He, Zhang, Huang, Zhang, Shu, Tao, She, Yu, et~al.]{wang2025mint}
Yi~Wang, Mushui Liu, Wanggui He, Longxiang Zhang, Ziwei Huang, Guanghao Zhang, Fangxun Shu, Zhong Tao, Dong She, Zhelun Yu, et~al.
\newblock Mint: Multi-modal chain of thought in unified generative models for enhanced image generation.
\newblock \emph{arXiv preprint arXiv:2503.01298}, 2025.

\bibitem[Wang et~al.(2024)Wang, Li, Li, and Liu]{wang2024genartist}
Zhenyu Wang, Aoxue Li, Zhenguo Li, and Xihui Liu.
\newblock Genartist: Multimodal llm as an agent for unified image generation and editing.
\newblock \emph{Advances in Neural Information Processing Systems}, 37:\penalty0 128374--128395, 2024.

\bibitem[Wei et~al.(2022)Wei, Wang, Schuurmans, Bosma, Xia, Chi, Le, Zhou, et~al.]{CoT_wei_2022}
Jason Wei, Xuezhi Wang, Dale Schuurmans, Maarten Bosma, Fei Xia, Ed~Chi, Quoc~V Le, Denny Zhou, et~al.
\newblock Chain-of-thought prompting elicits reasoning in large language models.
\newblock \emph{Advances in neural information processing systems (NeurIPS)}, 35:\penalty0 24824--24837, 2022.

\bibitem[Wu et~al.(2025{\natexlab{a}})Wu, Li, Zhou, Lin, Gao, Yan, Yin, Bai, Xu, Chen, et~al.]{QwenImage_wu_2025}
Chenfei Wu, Jiahao Li, Jingren Zhou, Junyang Lin, Kaiyuan Gao, Kun Yan, Sheng-ming Yin, Shuai Bai, Xiao Xu, Yilei Chen, et~al.
\newblock Qwen-image technical report.
\newblock \emph{arXiv preprint arXiv:2508.02324}, 2025{\natexlab{a}}.

\bibitem[Wu et~al.(2025{\natexlab{b}})Wu, Wang, Zhao, Yang, Zhang, Liu, Zhan, Han, Sun, Ji, et~al.]{reprompt_wu_2025}
Mingrui Wu, Lu~Wang, Pu~Zhao, Fangkai Yang, Jianjin Zhang, Jianfeng Liu, Yuefeng Zhan, Weihao Han, Hao Sun, Jiayi Ji, et~al.
\newblock Reprompt: Reasoning-augmented reprompting for text-to-image generation via reinforcement learning.
\newblock \emph{arXiv preprint arXiv:2505.17540}, 2025{\natexlab{b}}.

\bibitem[Wu et~al.(2024)Wu, Lian, Gonzalez, Li, and Darrell]{wu2024self}
Tsung-Han Wu, Long Lian, Joseph~E Gonzalez, Boyi Li, and Trevor Darrell.
\newblock Self-correcting llm-controlled diffusion models.
\newblock In \emph{Proceedings of the IEEE/CVF Conference on Computer Vision and Pattern Recognition}, pp.\  6327--6336, 2024.

\bibitem[Wu et~al.(2023{\natexlab{a}})Wu, Hao, Sun, Chen, Zhu, Zhao, and Li]{HPSv2_Wu_2023}
Xiaoshi Wu, Yiming Hao, Keqiang Sun, Yixiong Chen, Feng Zhu, Rui Zhao, and Hongsheng Li.
\newblock Human preference score v2: A solid benchmark for evaluating human preferences of text-to-image synthesis.
\newblock \emph{arXiv preprint arXiv:2306.09341}, 2023{\natexlab{a}}.

\bibitem[Wu et~al.(2023{\natexlab{b}})Wu, Sun, Zhu, Zhao, and Li]{HPS_Wu_2023}
Xiaoshi Wu, Keqiang Sun, Feng Zhu, Rui Zhao, and Hongsheng Li.
\newblock Human preference score: Better aligning text-to-image models with human preference.
\newblock In \emph{Proceedings of the IEEE/CVF International Conference on Computer Vision (ICCV)}, pp.\  2096--2105, October 2023{\natexlab{b}}.

\bibitem[Xu et~al.(2023)Xu, Liu, Wu, Tong, Li, Ding, Tang, and Dong]{imagereward_xu_2023}
Jiazheng Xu, Xiao Liu, Yuchen Wu, Yuxuan Tong, Qinkai Li, Ming Ding, Jie Tang, and Yuxiao Dong.
\newblock Imagereward: Learning and evaluating human preferences for text-to-image generation.
\newblock \emph{Advances in Neural Information Processing Systems (NeurIPS)}, 36:\penalty0 15903--15935, 2023.

\bibitem[Yang et~al.(2024{\natexlab{a}})Yang, Yu, Meng, Xu, Ermon, and Cui]{yang2024mastering}
Ling Yang, Zhaochen Yu, Chenlin Meng, Minkai Xu, Stefano Ermon, and Bin Cui.
\newblock Mastering text-to-image diffusion: Recaptioning, planning, and generating with multimodal llms.
\newblock In \emph{Forty-first International Conference on Machine Learning}, 2024{\natexlab{a}}.

\bibitem[Yang et~al.(2024{\natexlab{b}})Yang, Wang, Li, Lin, Lin, Liu, and Wang]{yang2024idea2img}
Zhengyuan Yang, Jianfeng Wang, Linjie Li, Kevin Lin, Chung-Ching Lin, Zicheng Liu, and Lijuan Wang.
\newblock Idea2img: Iterative self-refinement with gpt-4v for automatic image design and generation.
\newblock In \emph{European Conference on Computer Vision}, pp.\  167--184. Springer, 2024{\natexlab{b}}.

\bibitem[Zhang et~al.(2025)Zhang, Li, and Tai]{zhang2025layercraft}
Yuyao Zhang, Jinghao Li, and Yu-Wing Tai.
\newblock Layercraft: Enhancing text-to-image generation with cot reasoning and layered object integration.
\newblock \emph{arXiv preprint arXiv:2504.00010}, 2025.

\end{thebibliography}
